\documentclass{bmvc2k}

\usepackage{graphicx,booktabs,tabularx,xcolor,hyperref}
\usepackage{amsmath,amsthm,subcaption,float,amsfonts,amssymb,bm,mathtools,booktabs,threeparttable,nccmath}

\usepackage{microtype}
\usepackage{multirow}
\usepackage{cleveref}
\usepackage[title]{appendix}

\newtheorem{corollary}{Corollary}
\newtheorem{proposition}{Proposition}

\title{QINA: Quantum-Inspired Nonlinear Adapters for Pretrained Vision Models}
\addauthor{Mostafa Mehdipour Ghazi}{ghazi@di.ku.dk}{1}
\addinstitution{Pioneer Centre for AI\\University of Copenhagen}
\runninghead{M. Mehdipour Ghazi}{QINA}

\begin{document}

\maketitle

\begin{abstract}

Adapting large pretrained vision models under limited data and frozen-backbone constraints remains a central challenge in transfer learning. While lightweight adapters and parameter-efficient fine-tuning methods are widely adopted, most rely on generic multilayer perceptrons or low-rank linear updates, offering limited control over the spectral and geometric structure of feature transformations. We investigate whether structured nonlinear feature lifting can improve representational alignment in frozen regimes. We introduce Quantum-Inspired Nonlinear Adapters (QINA), compact modules that perform learnable trigonometric feature lifting followed by bounded nonlinear aggregation. The design induces structured oscillatory basis functions with an explicit norm-dependent Lipschitz bound, enabling spectral reshaping of pretrained representations without increasing the receptive field or significantly expanding parameter count. Importantly, the method operates entirely within standard deep learning frameworks and does not require quantum hardware. Through systematic experiments across natural and medical imaging datasets, classification and segmentation tasks, multiple adapters and placements, and varying training budgets, we show that performance in frozen regimes is primarily representation-limited. Nonlinear lifting improves adaptation, and the proposed structured trigonometric formulation consistently outperforms identity baselines, fixed Fourier feature mappings, and parameter-matched baseline adapters. Within the evaluated frozen-backbone settings, structured spectral parameterization provides a more effective inductive bias than generic nonlinear adapters. This work highlights the importance of geometry- and spectrum-aware adaptation mechanisms for large pretrained vision models.

\end{abstract}

\section{Introduction}

Modern computer vision relies on large-scale pretrained convolutional and transformer backbones. These models encode rich hierarchical representations that transfer effectively across tasks. In practice, however, many downstream applications operate under limited annotation budgets, restricted compute, or distribution shift. Freezing pretrained backbones and adapting them via lightweight modules or parameter-efficient fine-tuning (PEFT) \cite{xu2026parameter} has therefore become a standard paradigm. Despite strong empirical performance, most adapters rely on generic multilayer perceptrons (MLPs), implemented as serial or parallel bottleneck adapters \cite{houlsby2019parameter,chen2022adaptformer}, or low-rank linear updates (LoRA) \cite{hu2022lora}, offering limited control over the spectral or geometric structure of the induced feature transformation. This raises a fundamental question: Under frozen-backbone constraints, can a compact, structured nonlinear transformation provide more effective representational alignment than generic adapters?

Sinusoidal and structured nonlinear parameterizations have been explored in both representation learning and parameter-efficient adaptation. SIREN \cite{sitzmann2020implicit} uses periodic activations to learn implicit continuous representations, while FourierFT \cite{gao2024parameter} parameterizes weight updates through learnable spectral coefficients. Sine-LoRA \cite{ji2025efficient} applies sinusoidal nonlinearities to low-rank updates, and LoRAN \cite{deng2025enhancing} and AuroRA \cite{dong2026aurora} introduce nonlinear transformations within LoRA. DiTASK \cite{mantri2025ditask} further explores structured parameter-efficient adaptation across downstream tasks. These works suggest that spectral structure and nonlinear transformations can improve compact adaptation beyond conventional linear or MLP-based parameterizations.

In this work, we introduce quantum-inspired nonlinear adapters (QINA). Rather than redesigning full architectures, we propose a compact trigonometric lifting block that can be inserted into pretrained vision backbones. The module performs a learnable sinusoidal feature expansion followed by bounded nonlinear aggregation, analogous to expectation-value mappings in parameterized quantum circuits (PQCs), while remaining fully classical and executable within standard deep learning frameworks. In contrast to the weight-space adaptation methods, QINA operates directly on intermediate representations of frozen pretrained backbones, using learnable trigonometric lifting as a feature-space adaptation mechanism.

Our design is motivated by three observations. First, pretrained backbones impose strong structural inductive biases that determine which abstraction levels align with downstream labels. Second, domain shift weakens this representation alignment, reducing the advantage of particular layers. Third, structured feature lifts (e.g., Fourier mapping, sinusoidal encoding) provide principled mechanisms for increasing expressive capacity in a controlled manner, enabling efficient approximation of functions with high-frequency components \cite{caraiman2013quantum,shi2022quantum}. However, fixed Fourier expansions lack task adaptivity, while generic MLP adapters lack spectral structure. We hypothesize that a learnable, structured trigonometric lifting may better reshape pretrained representations under frozen constraints.

We conduct a systematic evaluation across natural and medical imaging datasets, classification and segmentation tasks, varying training budgets, and multiple adapters and placements. 
By freezing the backbone and training only the inserted block and task head, we isolate the representational contribution of the proposed transformation. We compare against identity (no adapter), fixed Fourier lifting, and parameter-matched MLP adapters (linear, ReLU, tanh, Houlsby \cite{houlsby2019parameter}, LoRA \cite{hu2022lora}). Our experiments reveal three consistent findings: (1) In frozen-backbone regimes, performance is primarily representation-limited rather than classifier-limited. (2) Structured nonlinear lifting improves adaptation, particularly under domain shift and low-data settings. (3) The proposed QINA module outperforms both fixed Fourier expansions and generic MLP adapters, suggesting that structured spectral parameterization provides a more effective inductive bias than unstructured nonlinear transformations.

Our main contributions are threefold: (1) We propose a compact quantum-inspired nonlinear adapter that performs learnable trigonometric feature lifting within pretrained vision backbones, enabling structured representation reshaping under frozen training. (2) We provide a theoretical perspective connecting the module to learnable Fourier feature expansions, spectral bias compensation, and controlled Lipschitz geometry. (3) Through systematic experiments across classification and segmentation tasks in natural and medical imaging, we demonstrate consistent improvements over identity, fixed Fourier, and parameter-matched MLP adapters, particularly in low-data and domain-shift regimes. Our objective is not to establish a universal PEFT leaderboard, but to isolate representation-space nonlinear adaptation under controlled frozen-backbone settings spanning CNN and ViT architectures, classification and segmentation, and natural-to-medical domain shifts.

\section{Related Work}

Quantum-based methods have increasingly been explored in computer vision and medical imaging \cite{yan2024review}. 
Although large-scale fault-tolerant quantum hardware remains unavailable, several works demonstrate that quantum formalism, tensor-network representations, or PQCs can be simulated and embedded within classical deep learning systems.

Existing methods fall into three categories. First, a number of works redesign segmentation architectures using quantum-inspired tensor representations or state-based neuron models \cite{konar2020quantum,konar20233,verma2025v3dqutrit}. 
These methods reinterpret pixel intensities or volumetric features as qubit or qutrit states and construct networks based on tensor-ring or state propagation mechanisms. While computationally efficient and fully executable on classical hardware, these approaches replace conventional architectures rather than integrate with modern pretrained models.

Second, segmentation has been formulated as a combinatorial optimization problem, solved via quantum annealing or simulated quantum circuits \cite{venkatesh2024q,domingo2024quantum}. Related quantum-inspired metaheuristics reinterpret classical optimization procedures such as genetic algorithms or particle swarm optimization through probabilistic quantum analogies \cite{li2017partitioned,das2018color,razmjooy2020skin}. 
These works primarily target optimization efficiency or alternative search dynamics, often requiring downsampled inputs due to hardware or simulation constraints, and do not address feature representation within pretrained deep networks.

Third, hybrid quantum–classical models insert small PQCs into convolutional or transformer-based architectures \cite{islam2025qpolypnet,venkatesh2024qubit,cherrat2024quantum,zhang2025hqvit,direkoglu2025quantum,telcceken2026new}. In these systems, simulated quantum circuits act as differentiable nonlinear modules within otherwise classical pipelines. Although promising, scalability remains constrained by circuit simulation overhead, and the representational role of such modules within frozen backbones has not been systematically isolated.

In contrast to prior work, we do not redesign architectures, formulate global optimization problems, or pursue quantum computational speedup. We study a controlled setting in which a compact quantum-inspired nonlinear adapter is inserted into pretrained backbones that remain frozen during adaptation. This design allows us to isolate the representational contribution of structured trigonometric feature lifting and bounded nonlinear dynamics, and to compare directly against alternative adapters under low-data and domain-shift regimes.

\section{Methods}

Conventional adapters for pretrained models typically rely on low-rank linear updates or generic MLPs. While effective, these mechanisms introduce largely unstructured nonlinear transformations. We instead consider frozen-backbone adaptation under limited data, where the geometry of pretrained representations should be preserved while enabling controlled feature reshaping. Our objective is to design a compact nonlinear adapter that introduces structured spectral lifting, maintains bounded activations, and remains computationally lightweight.

Let $\mathcal{F}_{\phi}$ denote a pretrained backbone with parameters $\phi$. Given downstream data $\mathcal{D} = \{(x_i, y_i)\}$, we operate in the frozen regime where $\phi$ remains fixed and only an adaptation module and task-specific head are optimized. We seek a structured nonlinear transformation that modulates intermediate features without disrupting pretrained representation geometry.

\begin{figure}[t]
\centering
\includegraphics[width=0.9\linewidth]{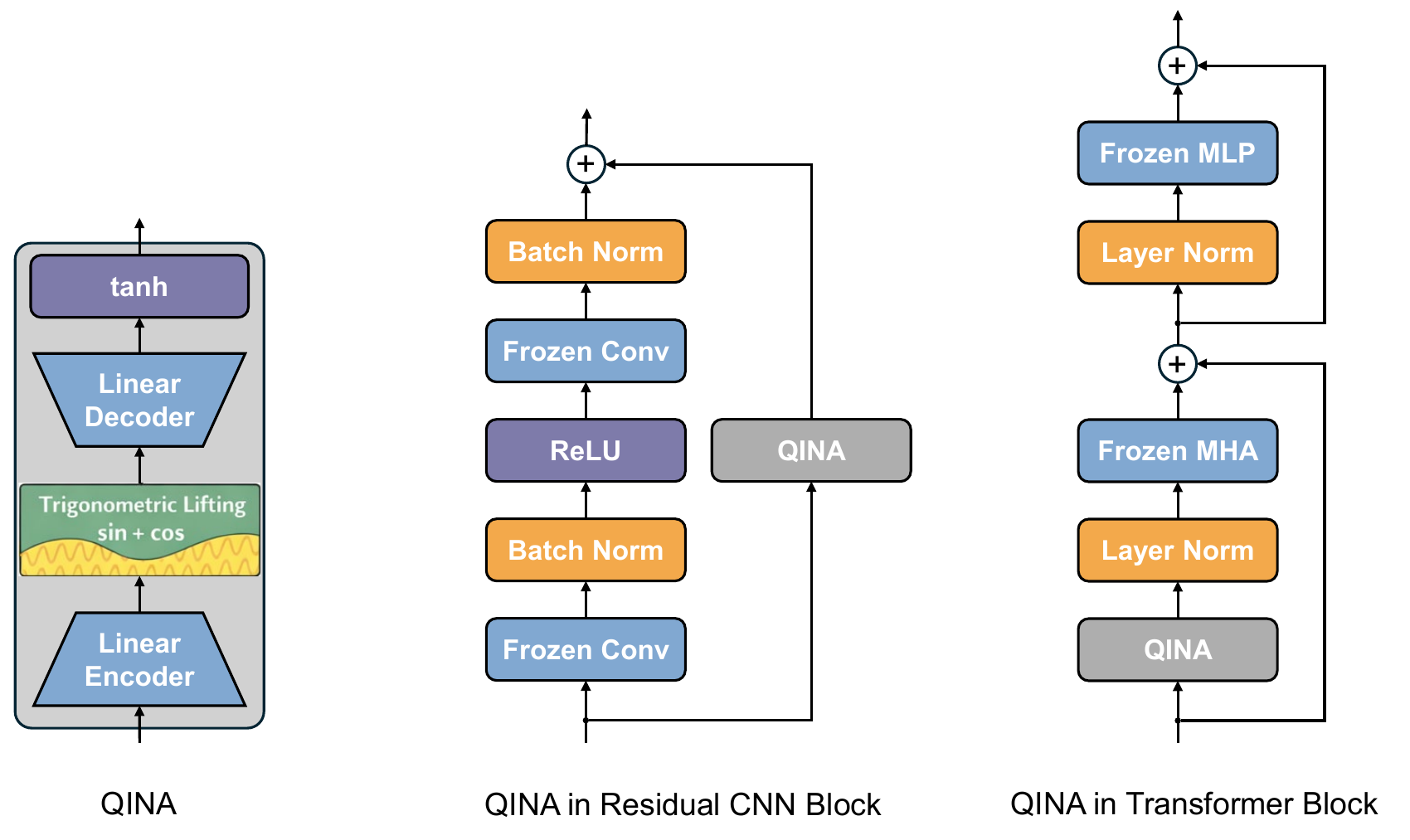}
\vspace{0.4cm}
\caption{Structured spectral adaptation via QINA in frozen backbones. Left: Internal architecture of the QINA module, consisting of linear encoding, trigonometric lifting, linear decoding, and bounded nonlinearity. Middle: In residual CNN blocks, QINA operates within the residual branch on intermediate feature maps, producing a bounded correction that is added to the shortcut connection while keeping the backbone parameters frozen. Right: In Transformer encoder layers, QINA is inserted before MHA in the first residual block.}
\label{fig:qina-block}
\end{figure}

We refer to this module as the \emph{Quantum-Inspired Nonlinear Adapter (QINA)}. Inspired by quantum information theory \cite{nielsen2010quantum}, where high-dimensional systems remain stable through structured operators and constrained state evolution \cite{lorenz2021causal}, QINA acts on latent features while preserving pretrained backbones. Although the design is inspired by trigonometric phase encodings used in quantum circuit formulations, it is implemented entirely as a differentiable real-valued transformation on classical hardware. \Cref{fig:qina-block} shows the proposed QINA block and how it can be integrated into the CNN and transformer backbones.

\subsection{QINA Formulation}

Let $X \in \mathbb{R}^{H \times W \times C}$ denote a feature tensor from an intermediate backbone layer. Flattening spatial dimensions yields $X_{\text{flat}} \in \mathbb{R}^{C \times N}$ with $N = H \times W$. QINA performs structured trigonometric lifting or phase encoding followed by bounded nonlinear aggregation:

\paragraph{Linear Projection}

\begin{equation}
P = W_{\text{enc}} X_{\text{flat}} + b_{\text{enc}} \in \mathbb{R}^{d \times N},
\end{equation}

\paragraph{Trigonometric Lifting}

\begin{equation} \label{eq:phase}
A = \sin(P) + \cos(P) \in \mathbb{R}^{d \times N},
\end{equation}

\paragraph{Linear Decoding}

\begin{equation}
S = W_{\text{dec}} A + b_{\text{dec}} \in \mathbb{R}^{C \times N},
\end{equation}

\paragraph{Bounded Nonlinearity}

\begin{equation}
R = \tanh(S),
\end{equation}

\noindent where $W_{\text{enc}} \in \mathbb{R}^{d \times C}$, $b_{\text{enc}} \in \mathbb{R}^{d}$, $W_{\text{dec}} \in \mathbb{R}^{C \times d}$, and $b_{\text{dec}} \in \mathbb{R}^{C}$ are learnable parameters and $d$ denotes the projection dimension.

The phase mapping in \Cref{eq:phase} can be interpreted as a learnable Fourier-type feature expansion. Unlike random Fourier features, projection frequencies and decoding weights are optimized jointly with the downstream objective. This structure introduces oscillatory basis functions capable of representing higher-frequency components that may not be efficiently captured by linear adapters.

The $\tanh$ activation bounds the adapter output, providing a functional analogy to the bounded expectation-value outputs in PQCs. QINA does not implement quantum states, observables, entanglement, or a quantum algorithm; it is a trigonometric adapter motivated by PQC functional forms. Without the final nonlinearity, the module reduces to a shallow linear decoding of learned trigonometric features. The bounded activation limits output magnitude and derivative amplification, providing controlled feature modulation under frozen-backbone adaptation. Thus, QINA acts as a structured spectral adapter rather than a generic MLP.

\subsection{Approximation Perspective}

QINA can be viewed as a learnable trigonometric feature expansion. We provide a simple approximation argument to clarify its expressive capacity.

\begin{proposition}
Let $f : \mathbb{R}^C \rightarrow \mathbb{R}$ be a continuous function with bounded spectral support. For any $\varepsilon > 0$, there exists a projection dimension $d$ and parameters $(W_{\mathrm{enc}}, W_{\mathrm{dec}})$ such that a QINA module approximates $f$ uniformly on compact subsets of $\mathbb{R}^C$ with error at most $\varepsilon$.
\end{proposition}

\paragraph{Proof Sketch.}
By Fourier approximation, such functions admit finite trigonometric expansions:
\begin{equation}
f(x) \approx \sum_{k=1}^{d} \left( a_k \cos(\omega_k^\top x) + b_k \sin(\omega_k^\top x) \right).
\end{equation}
Choosing rows of $W_{\mathrm{enc}}$ as frequencies $\omega_k$ and letting $W_{\mathrm{dec}}$ implement the linear combination recovers this expansion. The applied $\tanh$ preserves universal approximation on compact domains while enforcing bounded output geometry. \qed

\begin{corollary}
Increasing the projection dimension reduces approximation error for functions with higher spectral complexity at fixed backbone representations.
\end{corollary}

\noindent This perspective explains the empirical observation that larger $d$ improves performance in low-data regimes: increasing $d$ increases effective spectral resolution without modifying the frozen backbone.

\paragraph{\textbf{Lipschitz Stability.}}

Let the transformation be defined as
\begin{equation}
\Phi(x) = \tanh\!\big(W_{\mathrm{dec}}(\sin(W_{\mathrm{enc}}x) + \cos(W_{\mathrm{enc}}x))\big).
\end{equation}

\noindent Using the chain rule, the Jacobian satisfies
\begin{equation}
\nabla \Phi(x) = \mathrm{diag}\!\left(1 - \tanh^2(\cdot)\right) \, W_{\mathrm{dec}} \, \mathrm{diag}\!\left(\cos(W_{\mathrm{enc}}x) - \sin(W_{\mathrm{enc}}x)\right) \, W_{\mathrm{enc}}.
\end{equation}

Since $|1-\tanh^2(\cdot)| \le 1$ and $|\cos(u) - \sin(u)| \le \sqrt{2}$, QINA is globally Lipschitz continuous with an input-independent but parameter-dependent bound:
\begin{equation}
\|\nabla \Phi(x)\|_2 \le \sqrt{2}\,\|W_{\mathrm{dec}}\|_2 \|W_{\mathrm{enc}}\|_2.
\end{equation}
Thus, the nonlinearities provide bounded input response, while the Lipschitz constant remains proportional to the learned spectral norms of $W_{\mathrm{enc}}$ and $W_{\mathrm{dec}}$. We do not impose explicit spectral-norm constraints, and this should not be seen as a norm-independent stability guarantee.

\paragraph{\textbf{Connection to Kernel Feature Maps.}}

Random Fourier features approximate shift-invariant kernels using fixed sampled frequencies. QINA can be interpreted as a learnable analogue in which projection frequencies are optimized jointly with the downstream task. This induces a data-dependent spectral feature map rather than a fixed random embedding.

\paragraph{\textbf{Relation to MLP Adapters.}}

Standard MLPs are universal approximators. QINA does not replace this result; instead, it restricts the hypothesis class to structured trigonometric expansions followed by bounded nonlinear aggregation. Compared to MLP adapters, this imposes an explicit spectral bias, improving stability and generalization when data is limited and the backbone is frozen.

\subsection{Integration into Vision Backbones}

\paragraph{CNN Backbones.}

In convolutional networks, QINA is inserted into the residual bottleneck structure, preserving residual learning behavior and allowing the residual correction branch to compensate if the nonlinear adapter introduces noise or limited expressivity. 
The adapter operates on intermediate feature maps and returns a correction of matching dimensionality, allowing the network to retain pretrained representations while enabling structured nonlinear modulation. Because the backbone remains frozen, insertion does not disrupt pretrained filters but provides a controlled residual adjustment.

% However, the main transformation branch contains rich, pretrained semantic filters, and inserting adapters directly into pretrained convolutional pathways risks catastrophic forgetting and destabilizing learned representations, especially in few-shot/frozen-backbone and low-data regimes.

\paragraph{Vision Transformers.}

In the transformer architecture, adapters are inserted before the multi-head attention (MHA) module within the first residual encoder block to better align with the attention mechanism. Empirical evaluation showed that placing the adapter after the MHA module, where it operates directly on the token embeddings, led to inferior performance.

% In transformers, there is no meaningful shortcut path in the same sense as residual CNNs. Instead, the residual addition is inline, and both paths are implicit in the computation.

\section{Experiments}

We empirically evaluate whether a compact, structured nonlinear adapter, inserted into a pretrained backbone, improves frozen-backbone adaptation under low-data and domain-shift settings. 
All experiments are designed to isolate representational effects from parameter count and architectural confounders.

\subsection{Training Protocol}

\paragraph{\textbf{Backbones.}}

For classification, we use pretrained ResNet-18 \cite{he2016deep}, Xception \cite{fran2017deep}, and ViT-Tiny \cite{dosovitskiyimage,touvron2021training}. ResNet-18 is used within DeepLabv3+ \cite{chen2018encoder} for segmentation. All backbones are pretrained on ImageNet using supervised objectives.

\paragraph{\textbf{Frozen Adaptation.}}

In the primary setting, all backbone parameters remain frozen. Only the inserted adapter and task-specific head are optimized. This isolates the effect of representation reshaping without modifying pretrained filters.

\paragraph{\textbf{Full Fine-Tuning.}}

For comparison, we also evaluate end-to-end fine-tuning, updating all backbone parameters. This assesses whether adapter advantages persist when representation geometry is allowed to change.

\paragraph{\textbf{Low-Data Regimes.}}

To evaluate representational efficiency, we simulate stratified labeling budgets of $\{5, 10, 20, 50, 100, 200, 500\}$ samples. A fixed 20\% held-out validation split is selected before subsampling and reused only for evaluation across all experiments. All budgeted training sets are drawn exclusively from the remaining 80\% of the data. All configurations follow the same fixed-step training protocol. Each configuration is repeated with five random seeds, resulting in 35 models per backbone per dataset.

\paragraph{\textbf{Optimization Setting.}}

We use Adam with an initial learning rate of $10^{-4}$. Batch size is 32 for classification and 8 for segmentation. Standard geometric and photometric augmentations (horizontal flip, rotation, translation, contrast/brightness jitter, Gaussian noise) are applied during training. To ensure fair comparison across budgets, we normalize training by fixing the total number of optimization steps to 500 per experiment. The number of epochs is then adjusted according to dataset size, batch size, and optimization steps, and bounded within [10,100] to prevent under- or overfitting.

\subsection{Datasets and Tasks}

\begin{figure}[t]
\begin{minipage}{0.13\textwidth}
\centering
\includegraphics[height=\linewidth,width=\linewidth]{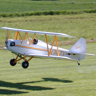}
\end{minipage}
\hfill
\begin{minipage}{0.13\textwidth}
\centering
\includegraphics[height=\linewidth,width=\linewidth]{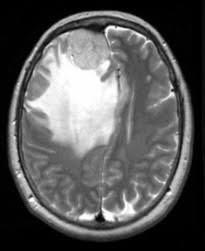}
\end{minipage}
\hfill
\begin{minipage}{0.13\textwidth}
\centering
\includegraphics[height=\linewidth,width=\linewidth]{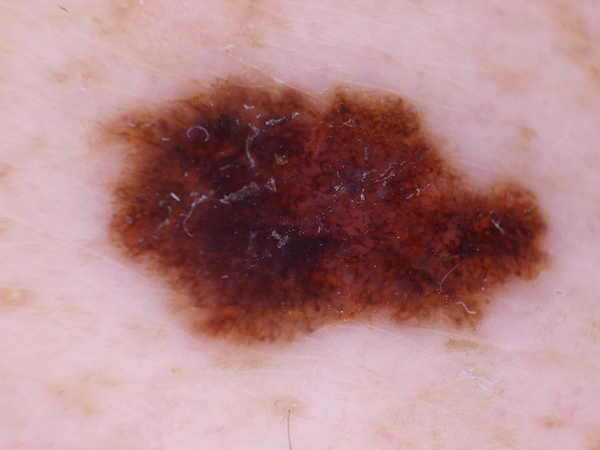}
\end{minipage}
\hfill
\begin{minipage}{0.13\textwidth}
\centering
\includegraphics[height=\linewidth,width=\linewidth]{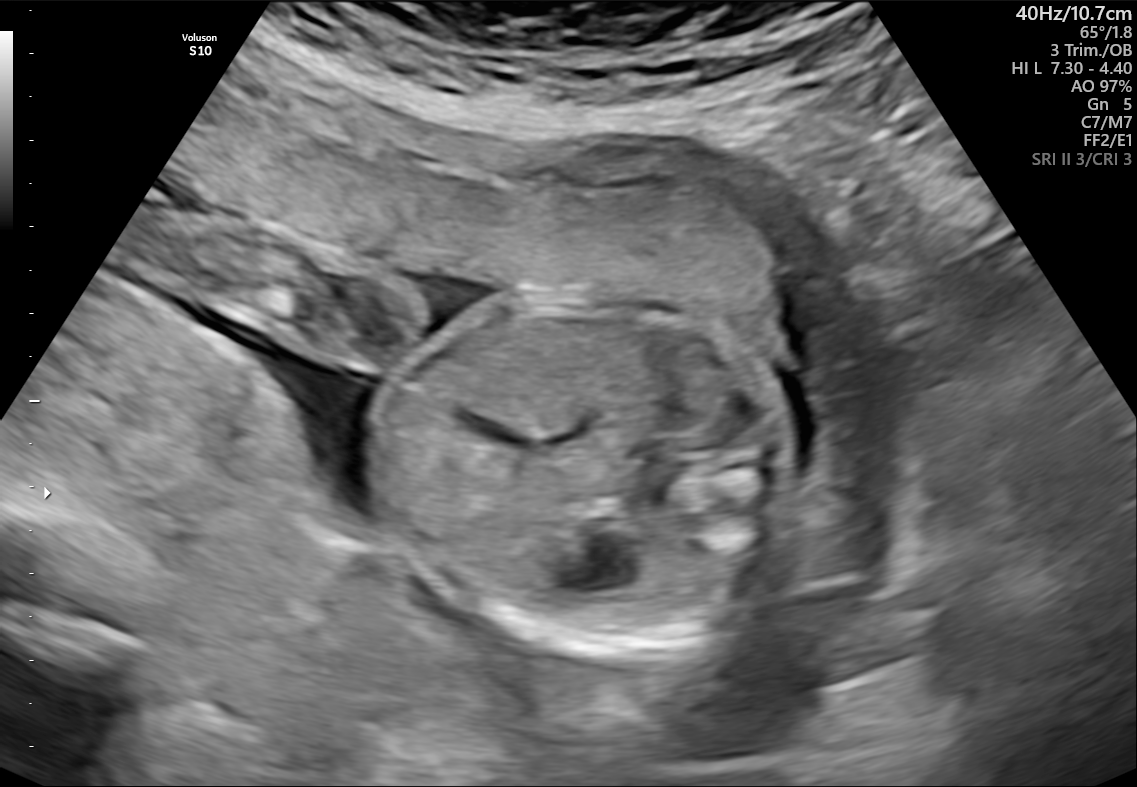}
\end{minipage}
\hfill
\begin{minipage}{0.13\textwidth}
\centering
\includegraphics[height=\linewidth,width=\linewidth]{fig_busi1.png}
\end{minipage}
\hfill
\begin{minipage}{0.13\textwidth}
\centering
\includegraphics[height=\linewidth,width=\linewidth]{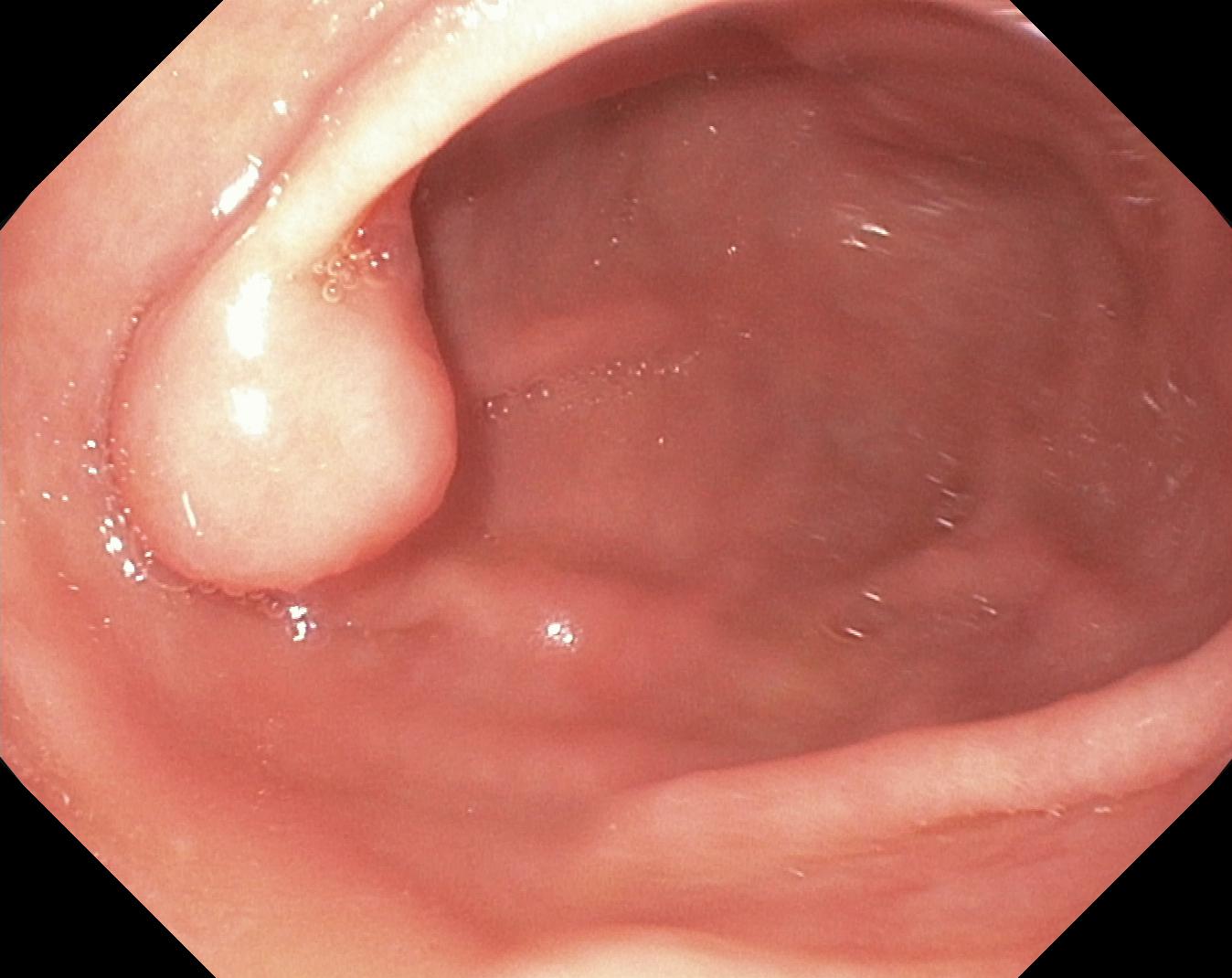}
\end{minipage}
\hfill
\begin{minipage}{0.13\textwidth}
\centering
\includegraphics[height=\linewidth,width=\linewidth]{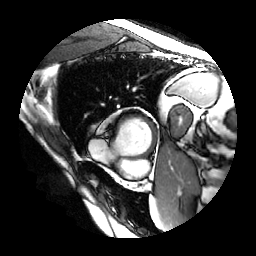}
\end{minipage}
\begin{minipage}{0.13\textwidth}
\centering
\includegraphics[height=\linewidth,width=\linewidth]{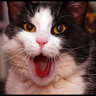}
\vspace{0.01cm}
\caption*{STL-10}
\end{minipage}
\hspace{0.005cm}
\begin{minipage}{0.13\textwidth}
\centering
\includegraphics[height=\linewidth,width=\linewidth]{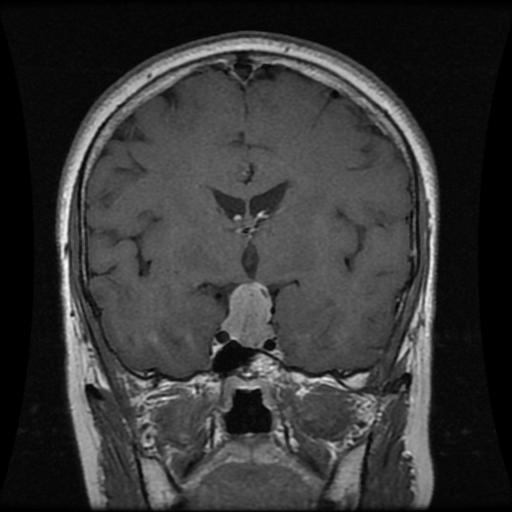}
\hfill
\caption*{BRISC}
\end{minipage}
\hspace{0.03cm}
\begin{minipage}{0.13\textwidth}
\centering
\includegraphics[height=\linewidth,width=\linewidth]{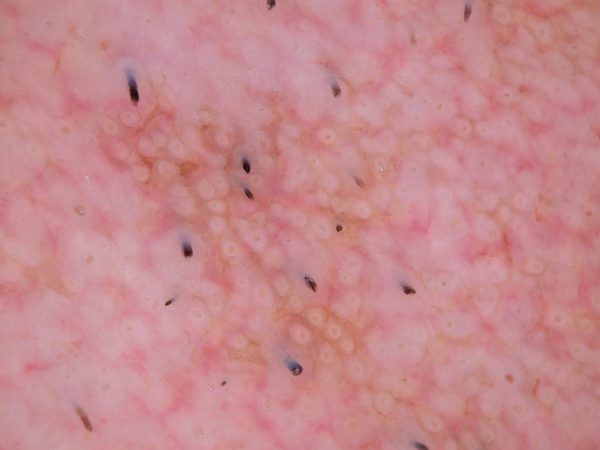}
\vspace{0.01cm}
\caption*{ISIC}
\end{minipage}
\hspace{0.005cm}
\begin{minipage}{0.13\textwidth}
\centering
\includegraphics[height=\linewidth,width=\linewidth]{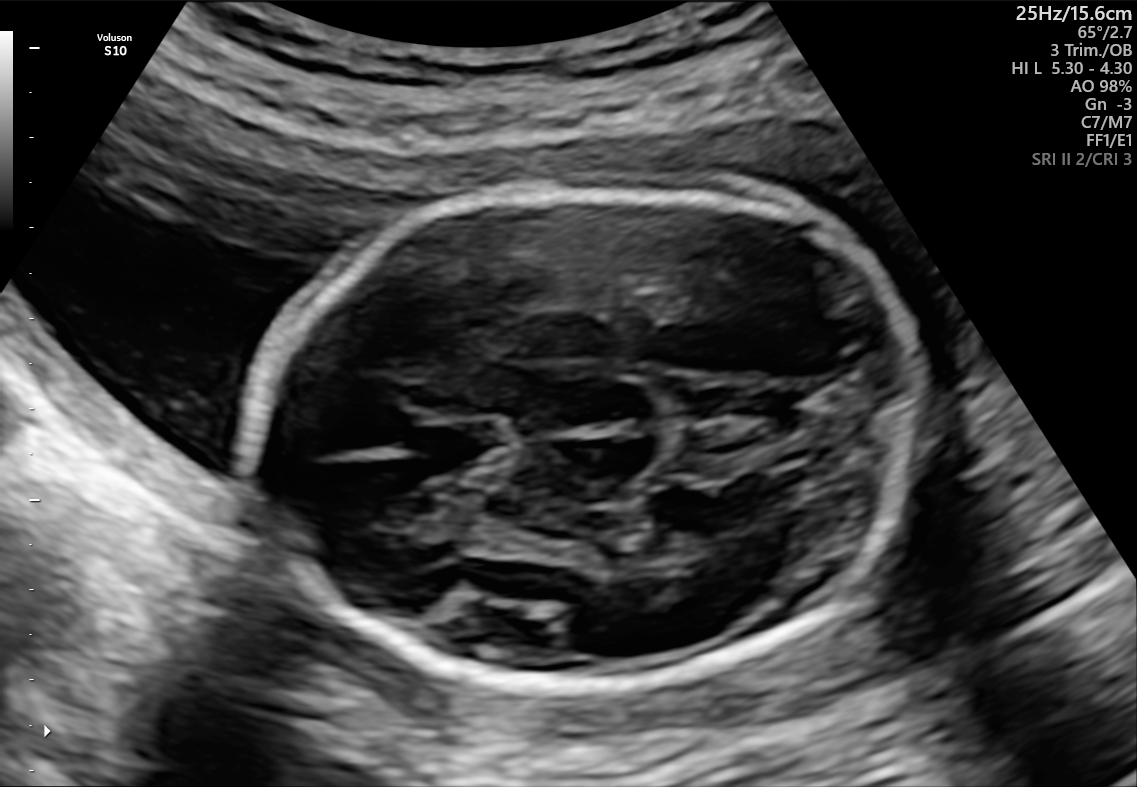}
\vspace{0.01cm}
\caption*{FETAL}
\end{minipage}
\hspace{0.05cm}
\begin{minipage}{0.13\textwidth}
\centering
\includegraphics[height=\linewidth,width=\linewidth]{fig_busi2.png}
\vspace{0.01cm}
\caption*{BUSI}
\end{minipage}
\hspace{0.04cm}
\begin{minipage}{0.13\textwidth}
\centering
\includegraphics[height=\linewidth,width=\linewidth]{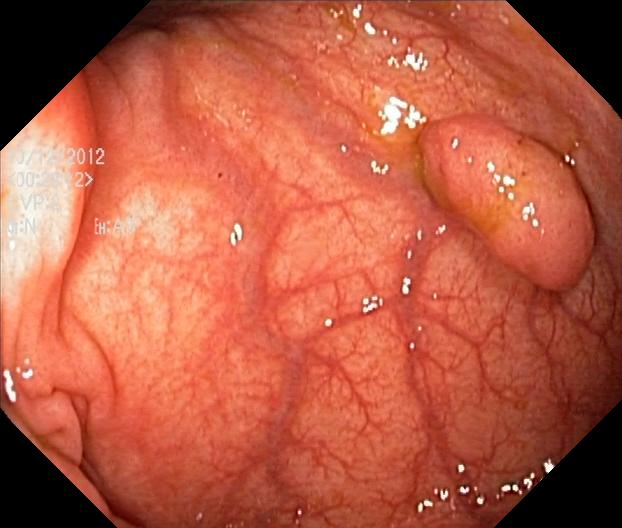}
\vspace{0.01cm}
\caption*{Kvasir}
\end{minipage}
\hspace{0.02cm}
\begin{minipage}{0.13\textwidth}
\centering
\includegraphics[height=\linewidth,width=\linewidth]{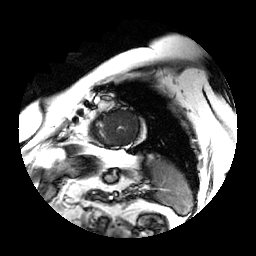}
\vspace{0.01cm}
\caption*{Cardiac}
\end{minipage}
\vspace{0.4cm}
\caption{Representative samples from natural and medical datasets. Natural images exhibit diverse color and semantic structure, whereas medical image modalities (MRI, ultrasound, dermoscopy, endoscopy) show distinct intensity statistics, texture patterns, and localized cues.}
\label{fig:sample-images}
\end{figure}

We evaluate the adapters across nine dataset-task settings from natural and medical imaging. Dataset statistics are summarized in \Cref{tab:dataset_summary} and representative samples are shown in \Cref{fig:sample-images}. Images are resized to $384 \times 384$ and intensities are scaled to the [0,255] range, followed by backbone-specific preprocessing.

\begin{table}[!b]
\caption{Summary of the datasets used in the study. The reported image resolutions correspond to average values due to variations across samples within each dataset.}
\vspace{0.4cm}
\label{tab:dataset_summary}
\centering
\renewcommand{\arraystretch}{1.15}
%\resizebox{0.999\textwidth}{!}{
\begin{tabular}{lccccc}
\toprule
\textbf{Dataset} & \textbf{Domain} & \textbf{Task} & \textbf{Dimension} &  \textbf{\# Class} & \textbf{\# Train} \\
\midrule
\textbf{STL-10} & Natural scene & Classification & 96$\times$96 & 10 & 5,000 \\
\midrule
\textbf{BRISC} & Brain tumor & Classification & 486$\times$475 & 4 & 5,000 \\
\textbf{ISIC} & Skin lesion & Classification & 905$\times$936 & 9 & 8,238 \\
\textbf{FETAL} & Fetal plane & Classification & 548$\times$822 & 6 & 7,129 \\
\midrule
\textbf{BUSI} & Breast tumor & Classification & 501$\times$616 & 3 & 780 \\
\midrule
\textbf{BRISC} & Brain tumor & Segmentation & 506$\times$505 & 2 & 3,933 \\
\textbf{BUSI} & Breast tumor & Segmentation & 495$\times$608 & 2 & 647 \\
\textbf{Kvasir} & Gastrointestinal polyp & Segmentation & 545$\times$625 & 2 & 1,000 \\
\textbf{Cardiac} & Left ventricle & Segmentation & 256$\times$256 & 2 & 1,610 \\
\bottomrule
\end{tabular}
%}
\end{table}

\paragraph{\textbf{Natural Images.}}

STL-10 \cite{coates2011analysis} is used for in-domain evaluation and ablation under controlled subsampling. It is derived from ImageNet, including color images of scenes and objects in 10 categories. While standardized benchmarks such as VTAB \cite{zhai2019large} provide broad task coverage, many constituent datasets share statistical properties with ImageNet pretraining, limiting their ability to probe representation alignment under genuine domain shift. We target medical imaging, which differs substantially from natural images in intensity statistics, texture characteristics, and semantic organization, reflecting the practical motivation for frozen-backbone adaptation: limited data and a significant domain gap.

\paragraph{\textbf{Medical Classification.}}

We evaluate on BRISC (brain MRI classification) \cite{fateh2026brisc}, ISIC (dermoscopic lesion classification) \cite{gessert2020skin}, and FETAL (fetal ultrasound plane classification) \cite{burgos2020evaluation}. To assess robustness under modality shift, we additionally evaluate transfer from FETAL-pretrained models to BUSI (breast ultrasound classification) data \cite{al2020dataset}.

\paragraph{\textbf{Medical Segmentation.}}

We evaluate segmentation on BRISC (brain tumor segmentation) \cite{fateh2026brisc}, BUSI (breast tumor segmentation) \cite{al2020dataset}, Kvasir (gastrointestinal polyp segmentation) \cite{jha2019kvasir}, and Sunnybrook Cardiac MRI (left ventricle segmentation) \cite{radau2009evaluation}. For segmentation experiments, only the pretrained backbone within DeepLabv3+ is modified; the decoder architecture remains unchanged.

\subsection{Baselines}

To ensure improvements are not attributable to parameter count or adapter insertion depth, we construct controlled baselines:

\paragraph{\textbf{Identity Block.}}

Frozen pretrained backbone with task-specific linear head only.

\paragraph{\textbf{MLP Adapter.}}

A parameter-matched MLP block inserted at the same location as QINA. We evaluate linear, ReLU, and tanh activations instead of phase encoding.

\paragraph{\textbf{Fourier Features.}}

A sinusoidal embedding $\phi(x) = \sin(Wx)$ with randomly sampled frequencies ($W \sim \mathcal{N}(0, \sigma^2)$). This baseline controls for gains arising from adding trigonometric basis functions without learnable spectral structure.

\paragraph{\textbf{QINA Block.}}

Learnable trigonometric projection with learned frequencies followed by bounded nonlinear aggregation. Adapters are inserted at all-, low-, mid-, and high-level feature layers. Projection dimension $d$ varies from 16 to 512 to analyze spectral capacity scaling and the impact of adapter dimensionality.

\paragraph{\textbf{LoRA Block.}}

A low-rank adapter \cite{hu2022lora,he2022towards} applied as an additive residual update to the network layers, using a trainable parameter budget of comparable scale to the other adapters.

\paragraph{\textbf{Houlsby Block.}}

A two-layer MLP adapter with GELU activations and skip connections \cite{houlsby2019parameter,he2022towards} applied at two locations within each network layer, with the number of trainable parameters matched to that of QINA for a fair comparison.

% \subsection{Evaluation Metrics}
% 
% For classification, we report accuracy and F1-score. For segmentation, we report Intersection-over-Union (IoU) and Dice score (F1-score) \cite{maier2024metrics}. We additionally report parameter counts and training time to quantify computational overhead.

\section{Results and Discussion}

Unless otherwise stated, we set the hidden dimension of the adapters to $d=128$ and insert them at the top layer (specifically, penultimate block) of each frozen backbone. This configuration is selected based on the ablation study in \Cref{ablation}, where it provides a strong balance between accuracy and computational complexity. For clarity of presentation, we visualize the main results in plots; detailed numerical results and complexity comparisons are provided in \textbf{Appendix}.

\subsection{Natural Image Classification}

\begin{figure}[t]
\begin{minipage}{0.32\textwidth}
\centering
%\caption*{A}
\includegraphics[width=\linewidth]{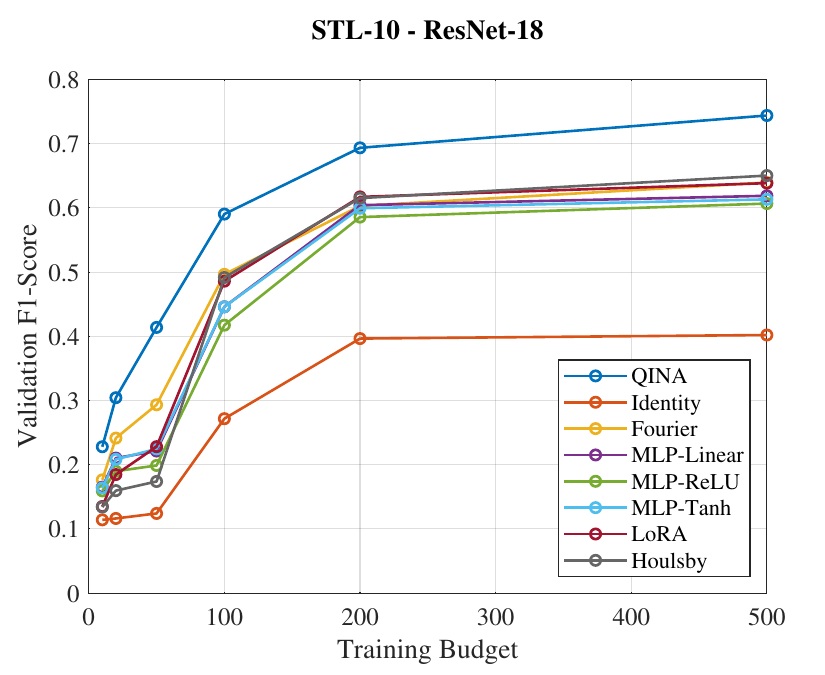}
\end{minipage}
\hfill
\begin{minipage}{0.32\textwidth}
\centering
%\caption*{B}
\includegraphics[width=\linewidth]{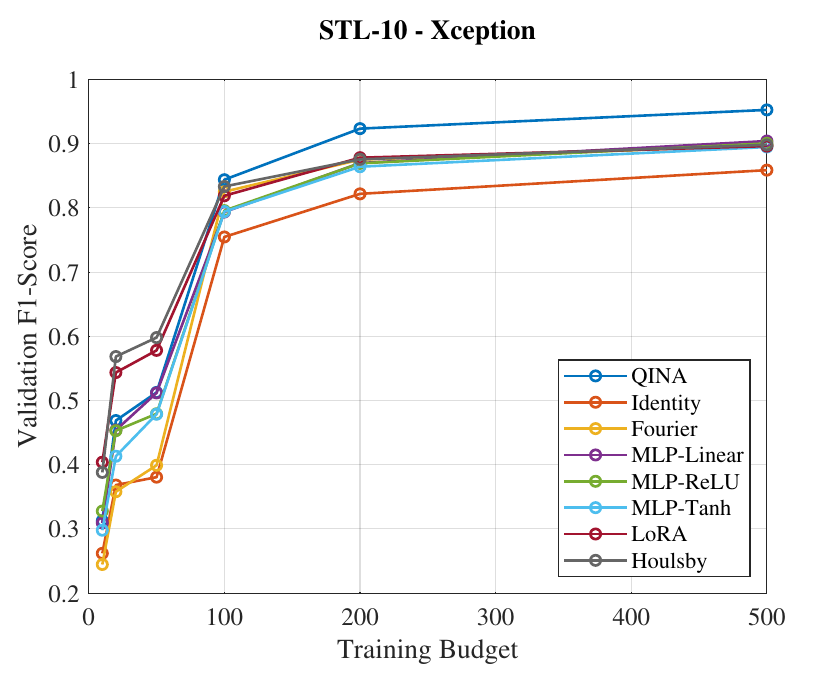}
\end{minipage}
\hfill
\begin{minipage}{0.32\textwidth}
\centering
%\caption*{C}
\includegraphics[width=\linewidth]{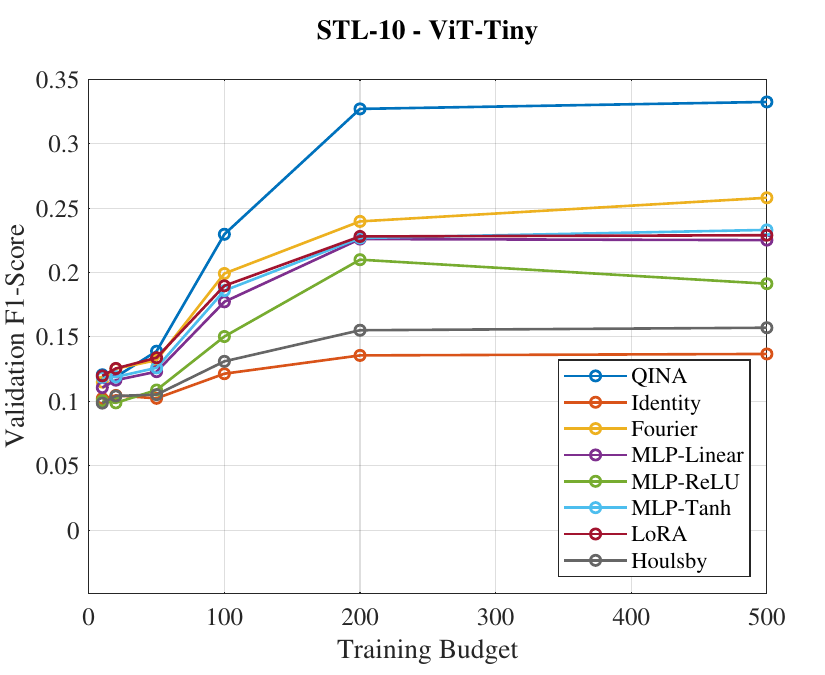}
\end{minipage}
\vspace{0.4cm}
\caption{Validation performance on STL-10 under frozen-backbone training across different classification backbones and training budgets. QINA consistently improves performance across models and budgets.}
\label{fig:stl-10_class_models}
\end{figure}

We first evaluate on STL-10 using different pretrained classification backbones. \Cref{fig:stl-10_class_models} summarizes model performance across training budgets. Across all budgets and architectures, inserting QINA consistently improves validation accuracy compared to the baselines. The improvement is particularly pronounced for ViT-Tiny at moderate-to-high budgets. Notably, the performance gap between QINA and the identity baseline  (frozen backbone with no adapter) remains substantial even as the number of labeled samples increases.

These results suggest that, under frozen-backbone training, performance is primarily representation-limited rather than head-capacity-limited. 
The linear head largely exploits the existing feature geometry, whereas QINA introduces a structured nonlinear transformation that better aligns pretrained features with the downstream label structure.

Importantly, the consistent superiority of QINA over both fixed Fourier expansions and generic MLP adapters suggests that performance gains cannot be attributed solely to increased nonlinear capacity. Instead, the structured spectral parameterization appears to provide a more effective mechanism for reshaping pretrained representations under frozen constraints.

\subsection{Medical Image Classification}

\begin{figure}[t]
\begin{minipage}{0.32\textwidth}
\centering
%\caption*{A}
\includegraphics[width=\linewidth]{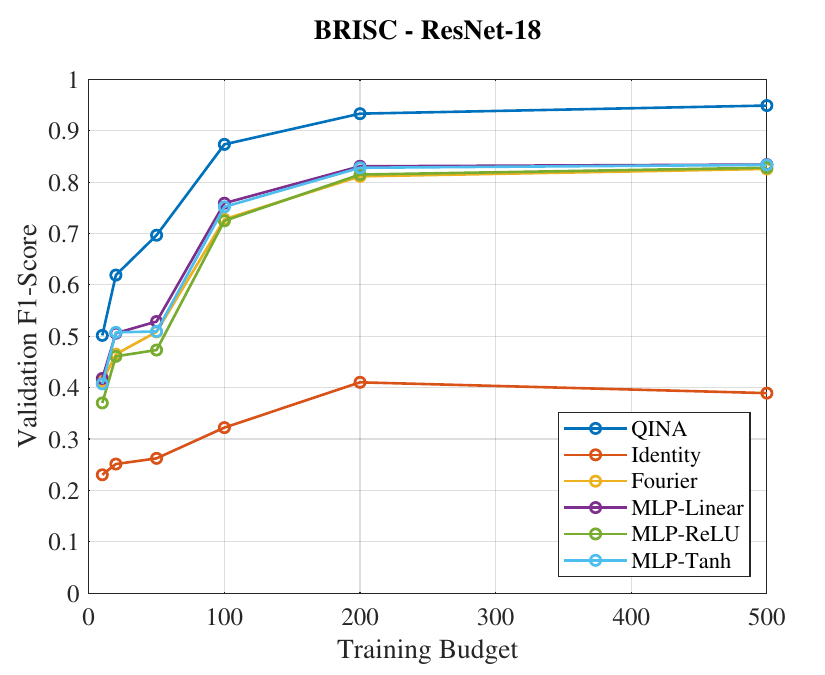}
\end{minipage}
\hfill
\begin{minipage}{0.32\textwidth}
\centering
%\caption*{B}
\includegraphics[width=\linewidth]{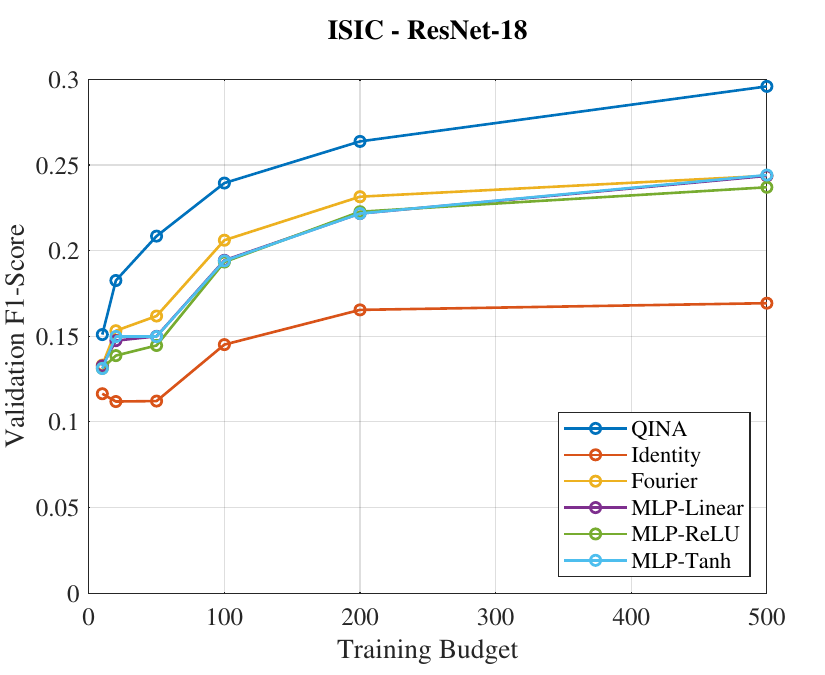}
\end{minipage}
\hfill
\begin{minipage}{0.32\textwidth}
\centering
%\caption*{C}
\includegraphics[width=\linewidth]{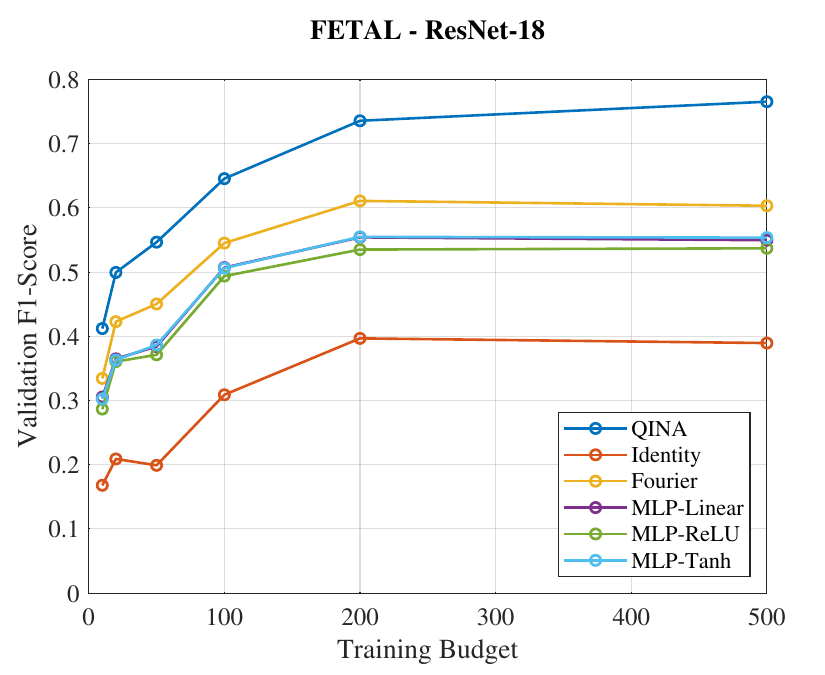}
\end{minipage}
\begin{minipage}{0.32\textwidth}
\centering
%\caption*{D}
\includegraphics[width=\linewidth]{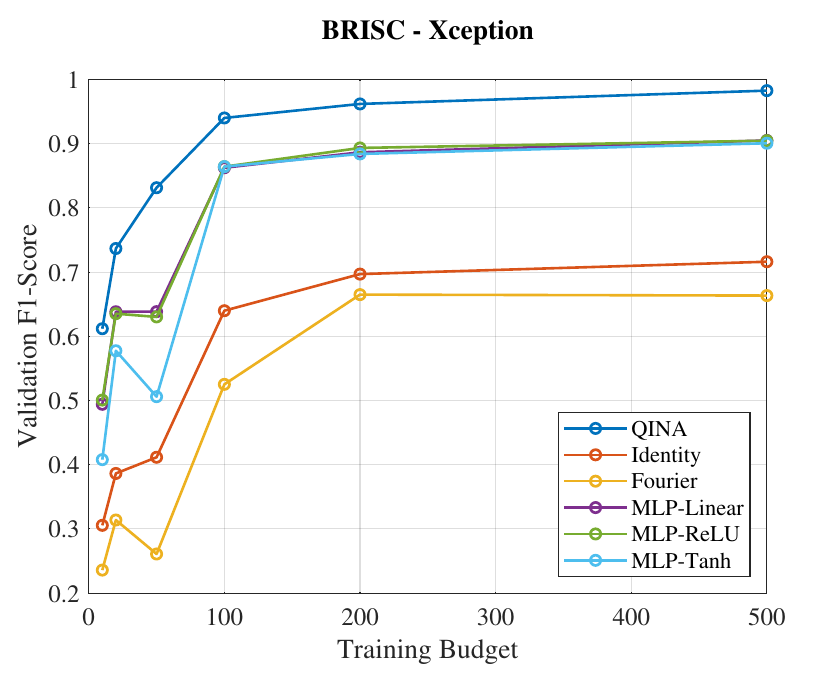}
\end{minipage}
\hfill
\begin{minipage}{0.32\textwidth}
\centering
%\caption*{E}
\includegraphics[width=\linewidth]{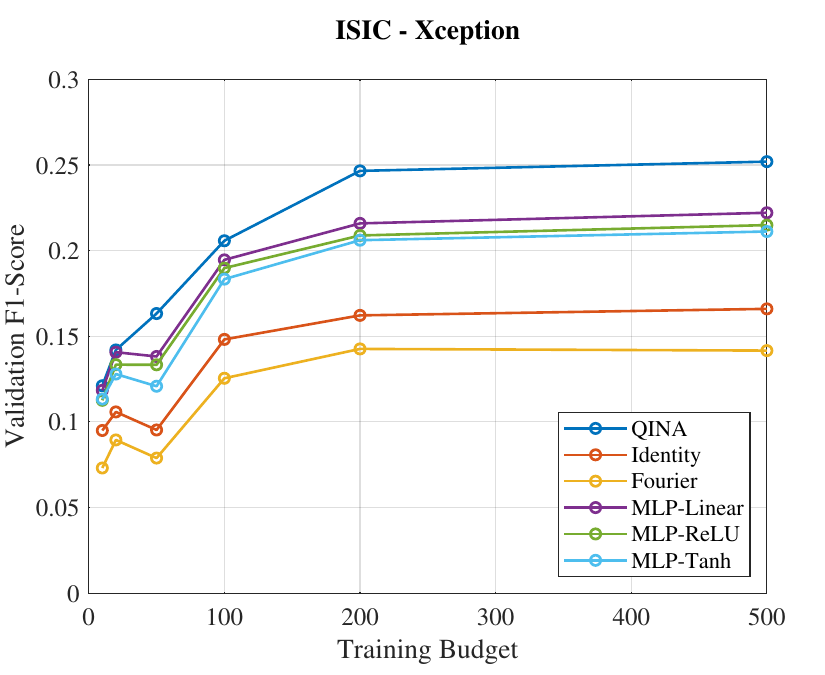}
\end{minipage}
\hfill
\begin{minipage}{0.32\textwidth}
\centering
%\caption*{F}
\includegraphics[width=\linewidth]{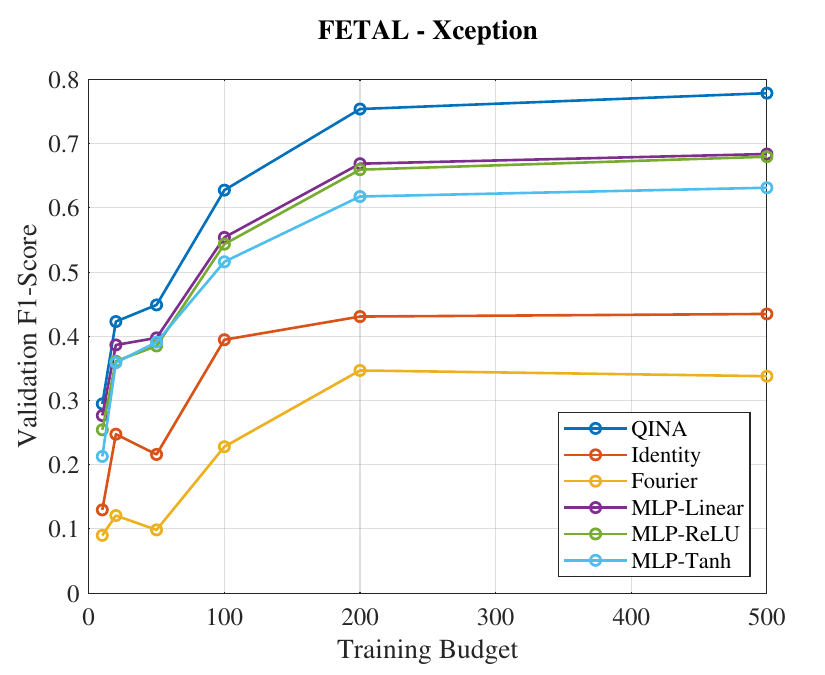}
\end{minipage}
\vspace{0.4cm}
\caption{Validation performance on medical classification datasets under frozen-backbone training across different CNN backbones and training budgets. QINA consistently improves performance across models, datasets, and budgets.}
\label{fig:medical_class_models}
\end{figure}

We next evaluate on medical classification datasets using pretrained CNN backbones. Results are shown in \Cref{fig:medical_class_models}. Again, QINA consistently improves performance across datasets and budgets. The relative gain is more pronounced for ResNet-18, particularly in lower-budget regimes. As in the natural-image setting, the identity baseline exhibits a clear performance gap compared to QINA.

The performance ranking across datasets and models remains stable as the training budget increases. This indicates that additional data improves decision boundary estimation but does not fundamentally alter which representation is most suitable. In other words, the dominant factor in frozen adaptation is representation alignment rather than classifier capacity or budget.

\paragraph{\textbf{Domain Shift Analysis.}}

We further analyze the role of domain shift. When pretrained models trained on natural images are applied to STL-10, semantic hierarchies and image statistics remain reasonably aligned with pretraining. In this regime, specific backbone layers exhibit clear abstraction advantages.

However, medical datasets differ substantially from natural images in image statistics (grayscale structure, modality-specific artifacts, low texture diversity) and semantic organization. Under such domain shift, no single backbone layer is strongly aligned with the downstream task. Early layers capture generic edges and noise patterns, while late layers encode high-level natural-image semantics that transfer weakly. Middle layers provide partial but imperfect alignment.

In this setting, improvements from QINA arise from nonlinear mixing and feature reshaping rather than intrinsic backbone quality. Additional labeled data refines the classifier but does not introduce missing invariances or modality-specific structure. For further interpretation, we fine-tune an ImageNet-pretrained ResNet on FETAL ultrasound data and subsequently freeze this adapted backbone for classification on BUSI. Although the tasks differ, the imaging modality is shared.

\begin{figure}[t]
\begin{minipage}{0.49\textwidth}
\centering
%\caption*{A}
\includegraphics[width=\linewidth]{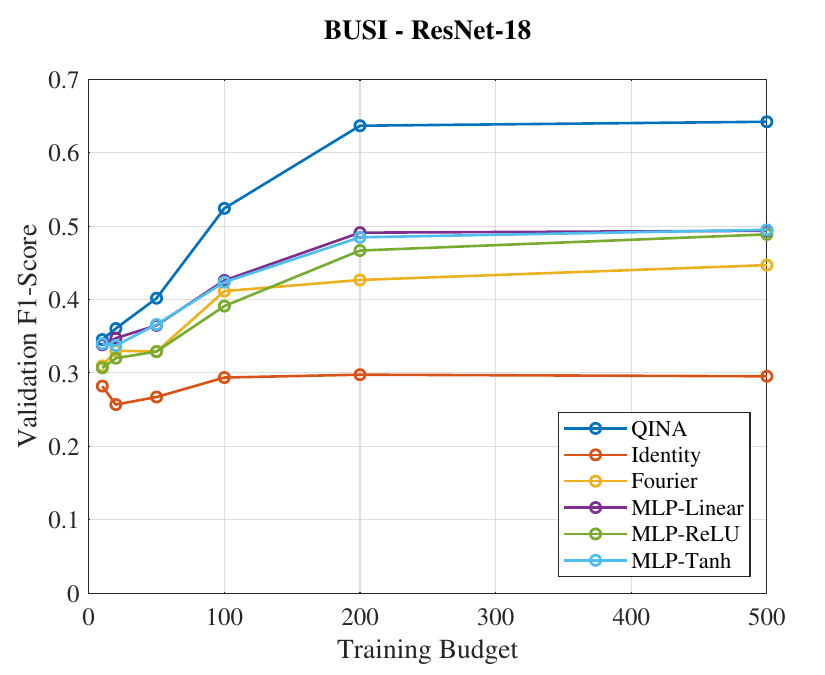}
\end{minipage}
\hfill
\begin{minipage}{0.49\textwidth}
\centering
%\caption*{B}
\includegraphics[width=\linewidth]{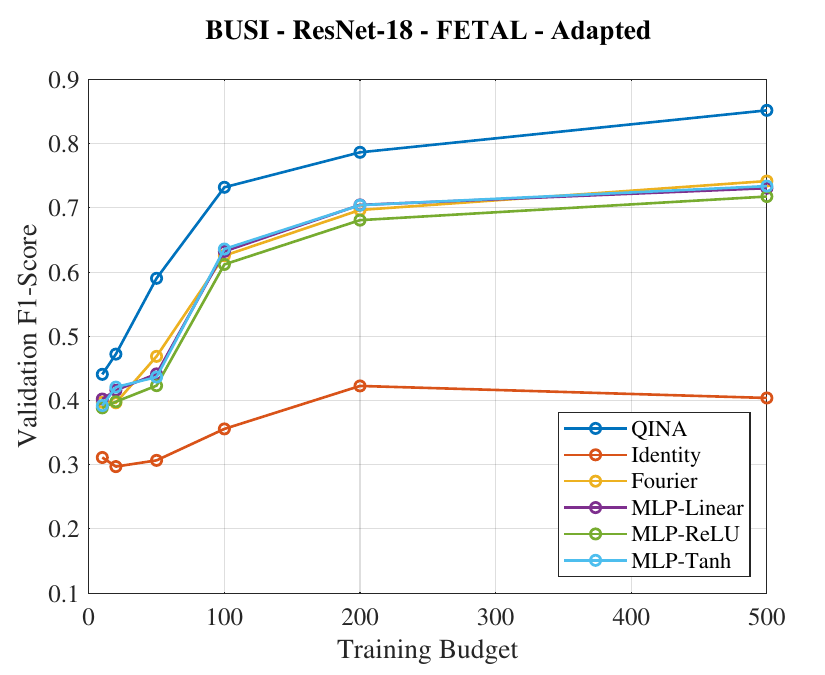}
\end{minipage}
\vspace{0.4cm}
\caption{BUSI classification performance using a frozen ImageNet-pretrained ResNet versus a frozen modality-adapted ResNet fine-tuned on FETAL ultrasound data. Although the tasks differ, modality alignment significantly improves downstream performance under freezing.}
\label{fig:busi_class_fetaltune}
\end{figure}

As shown in \Cref{fig:busi_class_fetaltune}, freezing the modality-adapted backbone yields a substantial performance improvement compared to freezing the original ImageNet backbone. After modality alignment, representation-level differences across layers re-emerge. Higher layers regain abstraction advantage, and the behavior becomes more similar to the STL-10 setting. This supports the interpretation that representation alignment, rather than head capacity, was the primary bottleneck under strong domain shift.

\subsection{Medical Image Segmentation}

\begin{figure}[t]
\begin{minipage}{0.49\textwidth}
\centering
%\caption*{A}
\includegraphics[width=\linewidth]{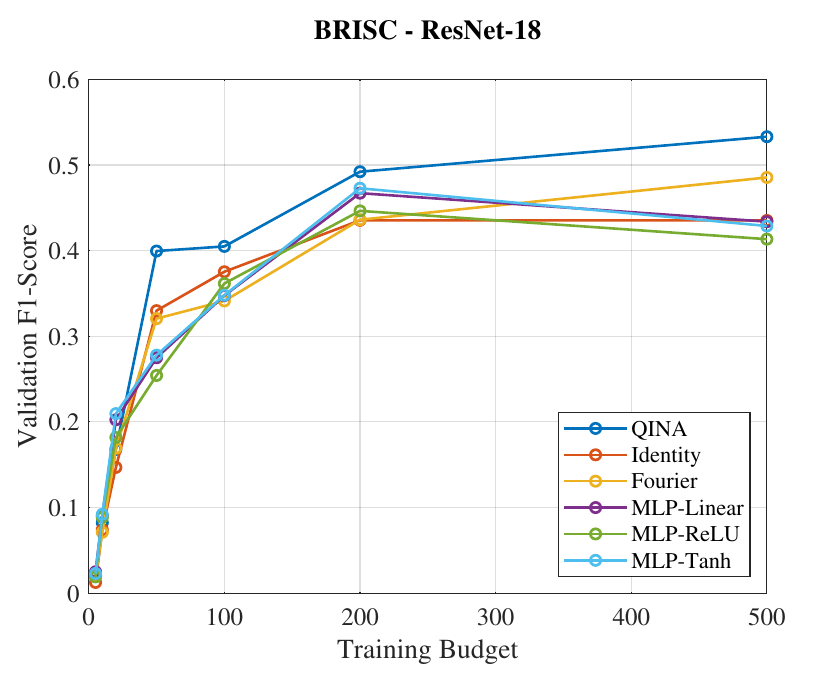}
\end{minipage}
\hfill
\begin{minipage}{0.49\textwidth}
\centering
%\caption*{B}
\includegraphics[width=\linewidth]{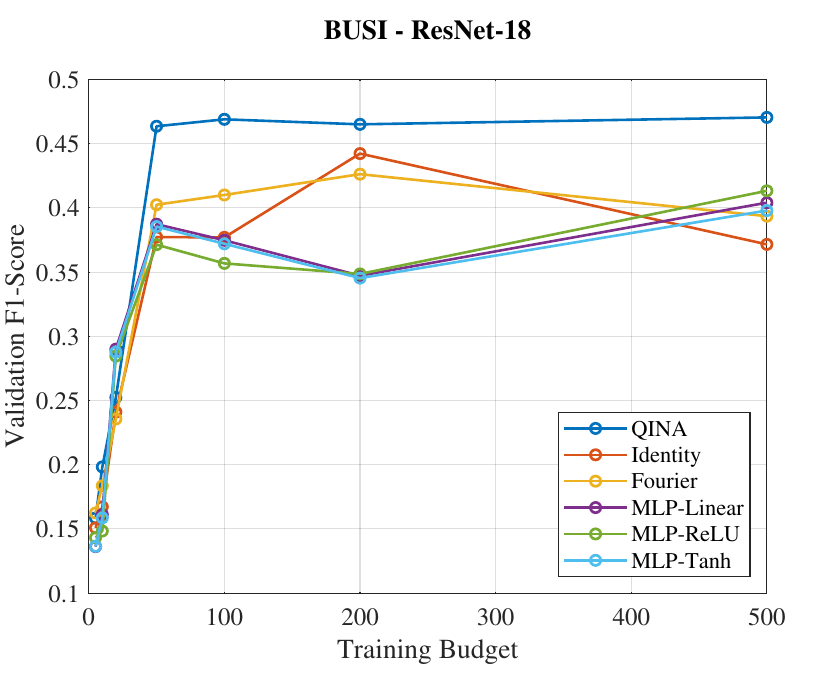}
\end{minipage}
\begin{minipage}{0.49\textwidth}
\centering
%\caption*{C}
\includegraphics[width=\linewidth]{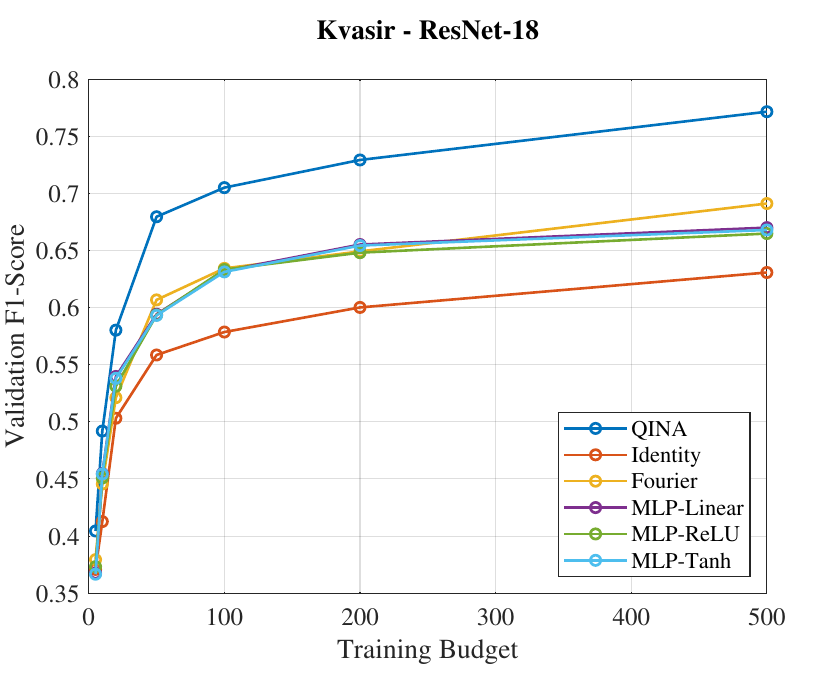}
\end{minipage}
\hfill
\begin{minipage}{0.49\textwidth}
\centering
%\caption*{D}
\includegraphics[width=\linewidth]{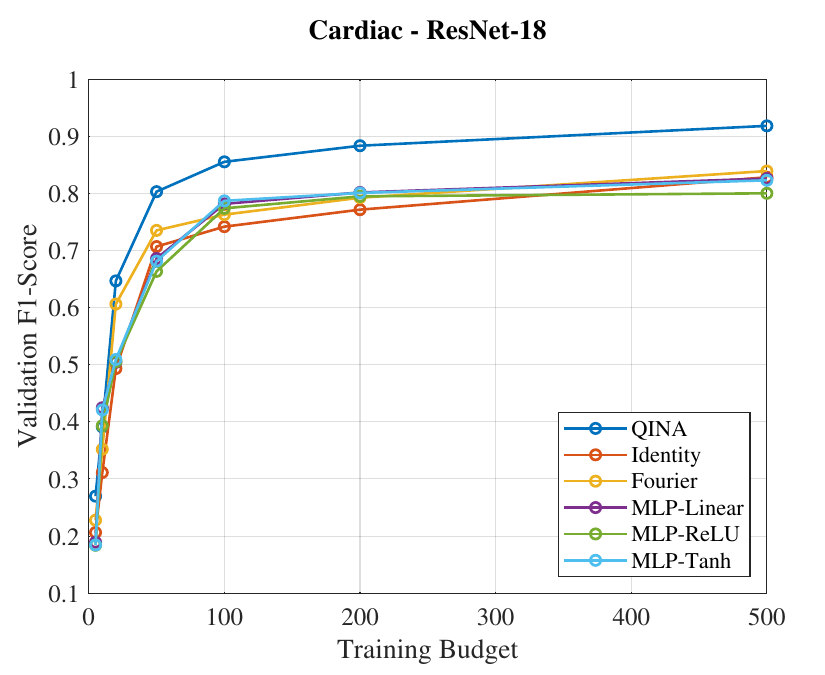}
\end{minipage}
\vspace{0.4cm}
\caption{Validation performance (foreground Dice) on medical segmentation datasets using DeepLabv3+ with a frozen pretrained encoder. Consistent improvements across budgets suggest that structured nonlinear adaptation benefits dense prediction tasks.}
\label{fig:medical_seg_models}
\end{figure}

To verify that the observed behavior is not specific to classification, we evaluate QINA within DeepLabv3+ segmentation pipelines while modifying only the pretrained backbone representation. As shown in \Cref{fig:medical_seg_models}, QINA improves validation performance across most budgets and datasets. Gains are particularly visible for Kvasir and Cardiac datasets at moderate-to-high budgets. The consistent gap relative to the identity baseline and MLP adapters indicates that structured nonlinear adaptation at the backbone level provides a more effective representation reshaping mechanism that benefits both classification and dense prediction tasks. These findings suggest that QINA enhances representational alignment rather than merely improving classifier expressivity.

\subsection{Ablation Study} \label{ablation}

We conduct controlled ablations on STL-10 to isolate three design factors: where adaptation is applied (placement), how much spectral capacity is provided (projection dimension), and whether the benefit is specific to frozen representations (full fine-tuning). \Cref{fig:stl-10_class_ablation} summarizes these experiments; complementary ablations are provided in \textbf{Appendix}.

\paragraph{\textbf{Placement.}}

Inserting QINA at higher layers consistently yields the strongest performance, followed by mid-layer and lower-layer placements. In contrast, distributing adapters across all layers with a matched total parameter budget is less effective, likely due to the introduction of low-dimensional bottlenecks. These results suggest that, under a frozen-backbone setting, adapting higher-level abstract representations with greater capacity is more beneficial than modifying early, edge-level features.

\paragraph{\textbf{Dimension.}}

Increasing the hidden dimension $d$ improves performance across budgets. The QINA block can be interpreted as a learnable Fourier-type feature expansion of dimension $d$. Larger $d$ increases effective feature rank and reduces approximation error of the nonlinear mapping, thereby enhancing representational capacity particularly in low-data regimes. However, computational cost also increases. We select $d=128$ as a balanced trade-off between accuracy and complexity across budgets.

\paragraph{\textbf{Fine-tuning.}}

When the backbone is unfrozen and trained end-to-end, the performance gap between QINA and baselines diminishes substantially. In this regime, optimization and inductive bias dominate, and data is no longer the bottleneck; the backbone adapts to the downstream task, reshaping the entire feature hierarchy. Fine-tuning erases layer-choice bias, and representational differences introduced by adapter placement become less critical.

This observation reinforces the primary conclusion of our study: QINA is particularly effective in frozen-backbone or low-data regimes where representation alignment is constrained. Under full fine-tuning with sufficient data, optimization can compensate for representational misalignment, reducing the relative benefit of lightweight adapters.

\begin{figure}[t]
\begin{minipage}{0.32\textwidth}
\centering
%\caption*{A}
\includegraphics[width=\linewidth]{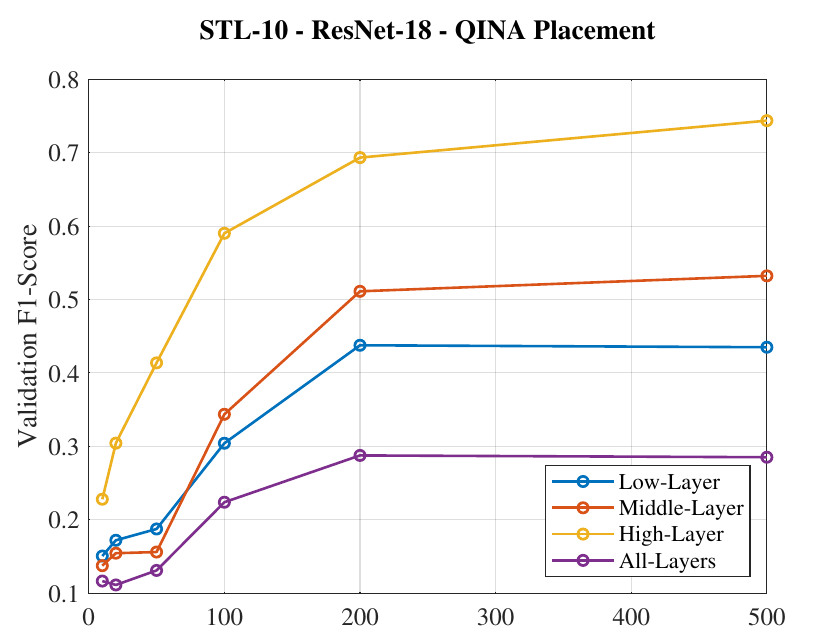}
\end{minipage}
\hfill
\begin{minipage}{0.32\textwidth}
\centering
%\caption*{B}
\includegraphics[width=\linewidth]{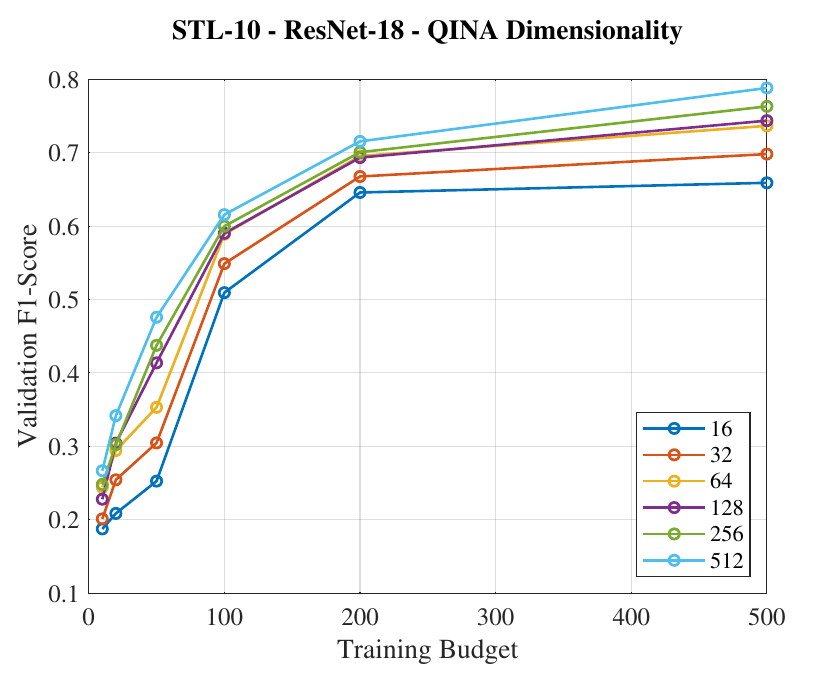}
\end{minipage}
\hfill
\begin{minipage}{0.32\textwidth}
\centering
%\caption*{C}
\includegraphics[width=\linewidth]{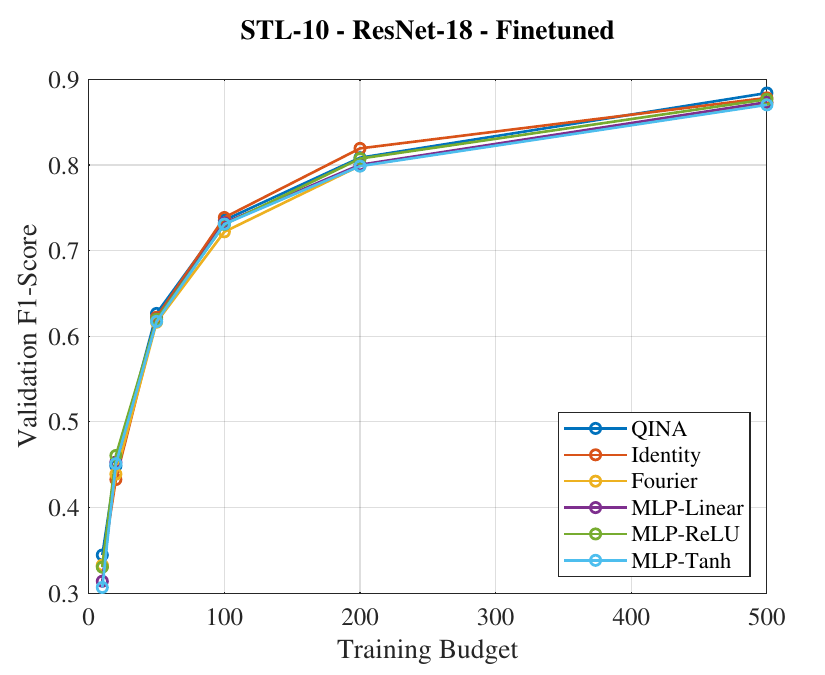}
\end{minipage}
\vspace{0.4cm}
\caption{Ablations on STL-10 evaluating QINA placement (left), hidden dimension (middle), and full fine-tuning (right). Higher-layer insertion and larger hidden dimension improve accuracy in the frozen regime, whereas end-to-end fine-tuning reduces adapter-specific gains.}
\label{fig:stl-10_class_ablation}
\end{figure}

\subsection{Representation Analysis}

QINA's trigonometric lifting introduces periodic basis functions that enable the adapter to represent higher-frequency components within the frozen backbone. Increasing the projection dimension $d$ expands the adapter's spectral resolution while leaving the backbone unchanged, providing a structured mechanism for reshaping feature geometry under frozen transfer.

To examine the resulting representations, we analyze feature statistics of a ResNet-18 backbone on STL-10 using a representative labeled budget of 500 training samples. Similar qualitative trends were observed across other labeling budgets. We evaluate three complementary metrics on the validation representations from the inserted blocks: Fisher Discriminant Ratio (FDR), which measures class separability; Effective Rank (ER) \cite{roy2007effective}, which characterizes the compactness of feature covariance; and Centered Kernel Alignment (CKA) \cite{kornblith2019similarity}, which quantifies similarity between adapted and baseline representations.

\begin{figure}[t]
\begin{minipage}{0.325\textwidth}
\centering
%\caption*{A}
\includegraphics[width=\linewidth]{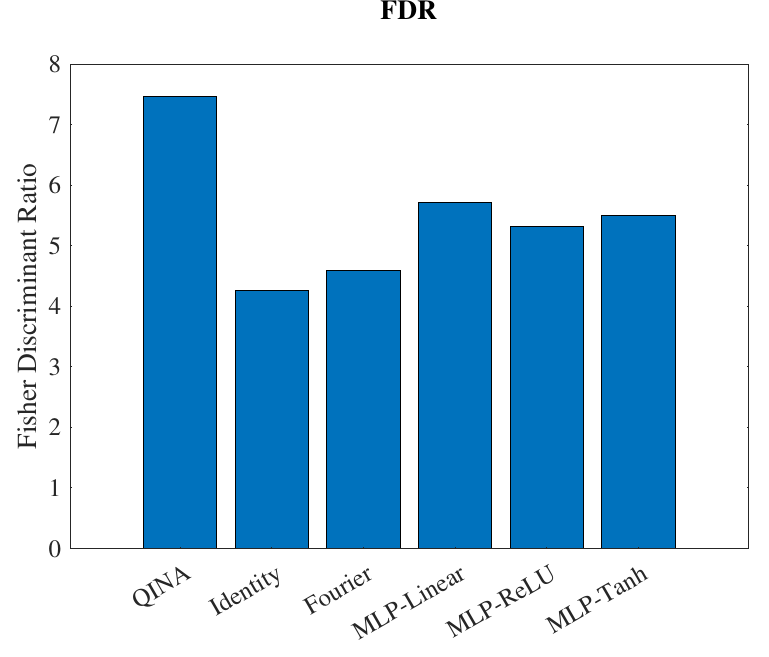}
\end{minipage}
\hfill
\begin{minipage}{0.325\textwidth}
\centering
%\caption*{B}
\includegraphics[width=\linewidth]{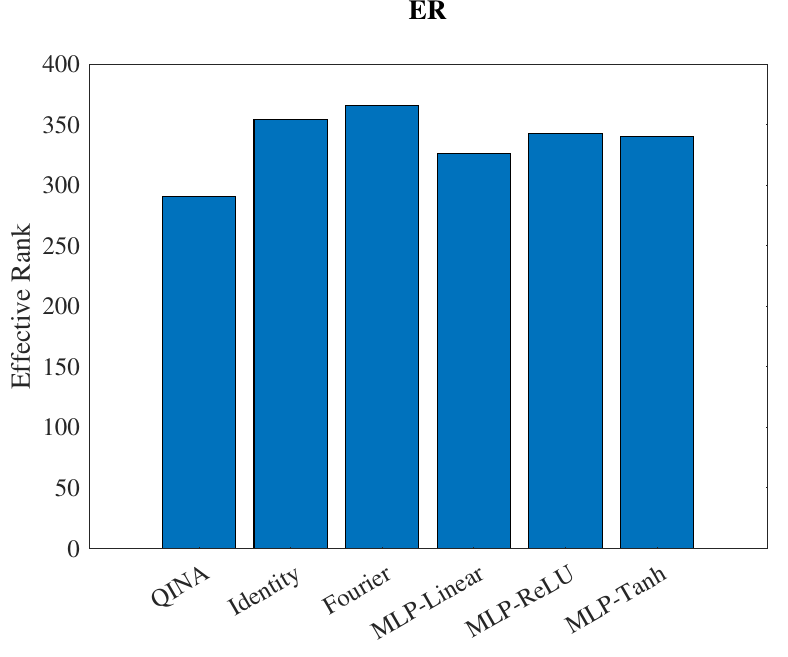}
\end{minipage}
\hfill
\begin{minipage}{0.3\textwidth}
\centering
%\caption*{C}
\includegraphics[width=\linewidth]{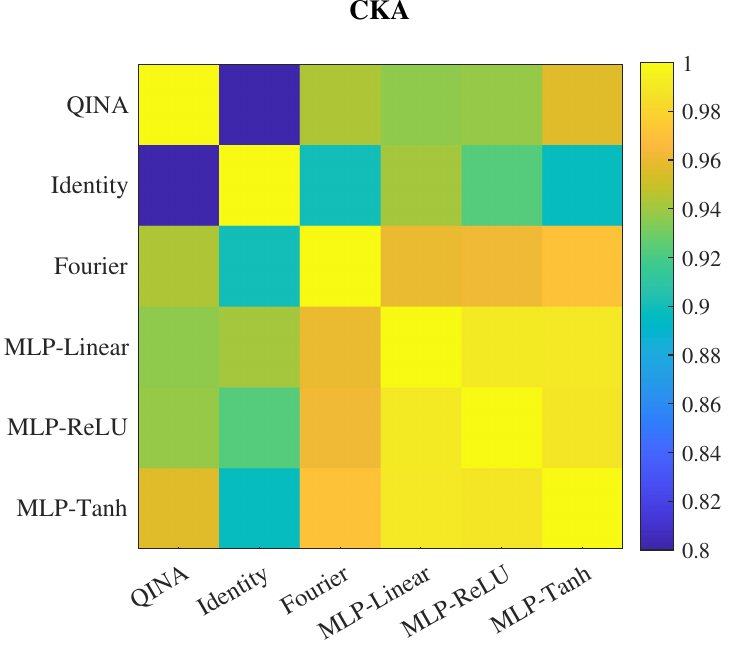}
\end{minipage}
\vspace{0.4cm}
\caption{Representation analysis of adapter features using a frozen ResNet-18 backbone on the STL-10 validation set. Adapters are compared using Fisher Discriminant Ratio (FDR), Effective Rank (ER), and Centered Kernel Alignment (CKA). QINA achieves higher FDR, indicating improved class separability, while exhibiting lower ER, suggesting more compact feature utilization. Lower CKA relative to the frozen backbone and alternative adapters indicates that QINA introduces a distinct transformation of the representation geometry.}
\label{fig:stl-10_class_represent}
\end{figure}

Results are shown in \Cref{fig:stl-10_class_represent}. Compared to identity, MLP, and Fourier baselines, QINA features achieve the highest FDR while exhibiting lower ER and lower CKA relative to the alternative blocks. Higher FDR indicates improved class separation, while reduced ER suggests more compact feature utilization. Lower CKA indicates that the adapter introduces a nontrivial transformation of the representation geometry rather than preserving the original backbone structure. Together, these observations suggest that QINA reshapes frozen feature representations toward more discriminative and compact embeddings while maintaining a low parameter footprint. This behavior is consistent with the role of structured trigonometric lifting in compensating for the low-frequency bias of frozen convolutional representations.

\section{Conclusion}

We investigated frozen-backbone adaptation through the lens of representation alignment and structured nonlinear transformation. Rather than redesigning full architectures or pursuing quantum computational speedup, we introduced a compact quantum-inspired nonlinear adapter (QINA) that performs learnable trigonometric feature lifting followed by bounded nonlinear aggregation.

QINA introduces three key properties: structured oscillatory basis functions applied within intermediate feature spaces, bounded smooth nonlinearities with a Lipschitz bound governed by the learned weight norms, and a lightweight residual formulation that preserves pretrained representation geometry. Together, these characteristics provide an inductive bias that complements frozen pretrained backbones without disrupting their hierarchical structure.

Across classification and segmentation tasks in natural and medical imaging, we observed three consistent findings. First, in frozen regimes, performance is primarily representation-limited rather than classifier-limited. Second, nonlinear lifting improves adaptation under low-data and domain-shift settings. Third, and most importantly, across the evaluated frozen-backbone settings, structured trigonometric lifting with learnable frequencies consistently outperforms both fixed Fourier embeddings and parameter-matched baseline adapters. This indicates that performance gains arise not merely from increased nonlinear capacity, but from the imposed spectral structure and controlled output geometry.

Our findings suggest that adapter effectiveness depends on representational alignment, placement, and geometric constraints. While quantum-inspired formulations motivate the trigonometric design, the proposed module operates entirely within standard deep learning frameworks and remains computationally efficient. Overall, this work highlights the importance of structured nonlinear adapters for frozen transfer learning and provides a controlled empirical and theoretical analysis of spectral feature lifting within modern vision backbones.

\subsection*{Acknowledgements}

This project is supported by the Pioneer Centre for AI, funded by the Danish National Research Foundation (grant number P1).

\newpage
\bibliography{references}

\newpage
\begin{appendices}

\section{PQC Variant}

In addition to the classical QINA block, we implement a hybrid quantum-classical variant in which the trigonometric lifting is replaced by a simulated parameterized quantum circuit (PQC). The PQC block first projects backbone features into a low-dimensional latent space, which is then encoded into a simulated quantum circuit through angle encoding. Expectation values of Pauli observables are used as nonlinear features and subsequently decoded back to the original channel dimension.

For each flattened spatial feature vector $x_i \in \mathbb{R}^{C}$ with $i = 1,\ldots,N$ flattened spatial positions, the projection step is defined as
\begin{equation}
p_i = W_{\text{enc}} x_i + b_{\text{enc}},
\end{equation}
\noindent where $p_i \in \mathbb{R}^{n}$ represents the encoded feature vector and $n$ denotes the number of qubits in the circuit. The projected features are then processed by a parameterized quantum circuit
\begin{equation}
a_i = f_{\theta}(p_i),
\end{equation}
\noindent where $f_{\theta}$ denotes a simulated quantum circuit with $n$ qubits, circuit depth $L$, and trainable rotation angles $\theta \in \mathbb{R}^{L \times n}$. Expectation values of Pauli-$Z$ observables are measured to produce nonlinear features $a_i \in \mathbb{R}^{n}$. Finally, the transformed features are decoded back into the backbone representation space:
\begin{equation}
R = W_{\text{dec}} A + b_{\text{dec}},
\end{equation}
\noindent where $A = [a_1, \ldots, a_N] \in \mathbb{R}^{n \times N}$ denotes the matrix of PQC-transformed features across spatial positions. Unlike the classical QINA block, no explicit $\tanh$ nonlinearity is required because expectation values of quantum measurements are inherently bounded in the interval $[-1,1]$. Both the circuit parameters $\theta$ and the classical encoding and decoding weights are optimized during training.

The PQC function $f_{\theta}$ operates on an $n$-qubit quantum state and produces nonlinear features through parameterized unitary transformations followed by expectation measurements, aligned with the structure of variational quantum circuits (VQCs). The circuit is initialized in the computational basis state
\begin{equation}
|\psi_0\rangle = |0\rangle^{\otimes n}.
\end{equation}

\paragraph{Data Encoding.}

The projected classical features $p_i = (p_{i1}, \ldots, p_{in})$ are encoded into the quantum state using single-qubit rotations around the Pauli-$X$ axis:
\begin{equation}
|\psi_{\text{enc}}\rangle = \left(\prod_{q=1}^{n} R_X(p_{iq})^{(q)}\right) |\psi_0\rangle,
\end{equation}
where $R_X(\alpha)^{(q)}$ denotes an $X$-rotation applied to qubit $q$ and
\begin{equation}
R_X(\alpha) =
\begin{bmatrix}
\cos(\alpha/2) & -i\sin(\alpha/2) \\
-i\sin(\alpha/2) & \cos(\alpha/2)
\end{bmatrix}.
\end{equation}

\paragraph{Parameterized Circuit Layers.}

The encoded state is processed by $L$ variational layers consisting of trainable single-qubit rotations followed by nearest-neighbor entangling gates. For layer $\ell$,
\begin{equation}
U_{\ell} = \left(\prod_{q=1}^{n-1} CZ_{q,q+1}\right) \left(\prod_{q=1}^{n} R_Y(\theta_{\ell,q})^{(q)}\right),
\end{equation}
where
\begin{equation}
R_Y(\alpha) =
\begin{bmatrix}
\cos(\alpha/2) & -\sin(\alpha/2) \\
\sin(\alpha/2) & \cos(\alpha/2)
\end{bmatrix}
\end{equation}
is a rotation around the Pauli-$Y$ axis with trainable angles $\theta=\{\theta_{\ell,q}\} \in \mathbb{R}^{L \times n}$, and $CZ_{q,q+1}$ denotes a controlled-$Z$ gate applied to neighboring qubits $q$ and $q+1$, defined as
\begin{equation}
CZ_{q,q+1} = I^{\otimes(q-1)} \otimes \text{diag}(1, 1, 1, -1) \otimes I^{\otimes(n-q-1)}.
\end{equation}
The overall circuit transformation is
\begin{equation}
|\psi_L\rangle = \left(\prod_{\ell=L}^{1} U_{\ell}\right) |\psi_{\text{enc}}\rangle .
\end{equation}

\paragraph{Expectation Measurement.}

Nonlinear features are obtained by measuring expectation values of Pauli-$Z$ operators for each qubit:
\begin{equation}
a_{iq} = \langle \psi_L | Z_q | \psi_L \rangle ,
\end{equation}
where
\begin{equation}
Z_q = I^{\otimes(q-1)} \otimes \text{diag}(1, -1) \otimes I^{\otimes(n-q)} .
\end{equation}
These expectation values satisfy $a_{iq} \in [-1,1]$. The resulting feature vector $a_i = (a_{i1}, \ldots, a_{in})$ forms the PQC-transformed representation used by the adapter.

\section{Gradient Computation}

\subsection{Backward Pass for QINA}

All operations in QINA are differentiable, and gradients are computed analytically using the chain rule. Using the derivative of the hyperbolic tangent, the gradient with respect to $S$ is
\begin{equation}
\frac{\partial \mathcal{L}}{\partial S} = \frac{\partial \mathcal{L}}{\partial R} \odot (1 - R^2),
\end{equation}
where $\odot$ denotes element-wise multiplication. For the linear decoder, the gradients are
\begin{align}
\label{eq:grad_dec1}
\frac{\partial \mathcal{L}}{\partial W_{\text{dec}}} = \frac{\partial \mathcal{L}}{\partial S} A^\top, \\
\label{eq:grad_dec2}
\frac{\partial \mathcal{L}}{\partial b_{\text{dec}}} = \sum_{i} \frac{\partial \mathcal{L}}{\partial S_i}, \\
\label{eq:grad_dec3}
\frac{\partial \mathcal{L}}{\partial A} = W_{\text{dec}}^\top \frac{\partial \mathcal{L}}{\partial S}.
\end{align}
The gradients for the phase encoding yield
\begin{equation}
\frac{\partial \mathcal{L}}{\partial P} = \frac{\partial \mathcal{L}}{\partial A} \odot (\cos(P) - \sin(P)).
\end{equation}
Finally, the gradients for the encoder projection become
\begin{align}
\label{eq:grad_enc1}
\frac{\partial \mathcal{L}}{\partial W_{\text{enc}}} = \frac{\partial \mathcal{L}}{\partial P} X^\top, \\
\label{eq:grad_enc2}
\frac{\partial \mathcal{L}}{\partial b_{\text{enc}}} = \sum_i \frac{\partial \mathcal{L}}{\partial P_i}, \\
\label{eq:grad_enc3}
\frac{\partial \mathcal{L}}{\partial X} = W_{\text{enc}}^\top \frac{\partial \mathcal{L}}{\partial P}.
\end{align}

\subsection{Backward Pass for PQC}

The PQC variant replaces the trigonometric lifting with a simulated parameterized quantum circuit. Since analytic derivatives of the simulated circuit were not implemented in our framework, gradients are estimated using numerical Jacobians based on central finite differences. However, the gradients for the adapter output follow the same form as in \Cref{eq:grad_dec1,eq:grad_dec2,eq:grad_dec3}, assuming that $R = S$.

For each spatial index $i$, the Jacobian $\partial A_i / \partial p_i$ is approximated using central finite differences. For feature dimension $k$,
\begin{equation}
\frac{\partial A_i}{\partial p_{ik}} \approx \frac{f_{\theta}(p_i + \epsilon e_k) - f_{\theta}(p_i - \epsilon e_k)}{2\epsilon},
\end{equation}
where $e_k$ denotes the $k$-th basis vector and $\epsilon$ is a small perturbation constant. The resulting gradient is
\begin{equation}
\frac{\partial \mathcal{L}}{\partial p_i} = \left(\frac{\partial A_i}{\partial p_i}\right)^\top \frac{\partial \mathcal{L}}{\partial A_i}.
\end{equation}
Gradients for the circuit parameters $\theta$ are computed similarly:
\begin{equation}
\frac{\partial A_i}{\partial \theta_j} \approx \frac{f_{\theta+\epsilon e_j}(p_i) - f_{\theta-\epsilon e_j}(p_i)}{2\epsilon}.
\end{equation}
Accumulating across spatial locations gives
\begin{equation}
\frac{\partial \mathcal{L}}{\partial \theta_j} = \sum_i \left(\frac{\partial A_i}{\partial \theta_j}\right)^\top \frac{\partial \mathcal{L}}{\partial A_i}.
\end{equation}
After computing the gradients with respect to the projected features $p_i$, the encoder gradients follow the same form as in \Cref{eq:grad_enc1,eq:grad_enc2,eq:grad_enc3}. Thus, the PQC adapter combines analytic gradients for the classical components with numerically estimated Jacobians for the quantum circuit simulation.

% For rotation gates of the form $R(\theta) = \exp(-i\theta G/2)$, with generator eigenvalues $\pm 1$, the exact gradient is:
% \begin{equation}
% \frac{\partial f}{\partial \theta} = \frac{1}{2} \left( f(\theta + \frac{\pi}{2}) - f(\theta - \frac{\pi}{2}) \right).
% \end{equation}
% This yields exact gradients for rotation gates ($R_X$, $R_Y$) without numerical instability. For a circuit of many parameters, each partial derivative can be computed w.r.t. $\theta_j$ by shifting that parameter by $\pm\pi/2$ while keeping others fixed and forming the difference. 

\section{QINA Component Ablations}

To isolate the contributions of QINA's trigonometric encoding and bounded output nonlinearity, we evaluate alternative choices on STL-10 using ResNet-18 while keeping the backbone, adapter placement, hidden dimension,
training budgets, and optimization protocol fixed.

\begin{figure}[t]
\begin{minipage}{0.49\textwidth}
\centering
%\caption*{A}
\includegraphics[width=\linewidth]{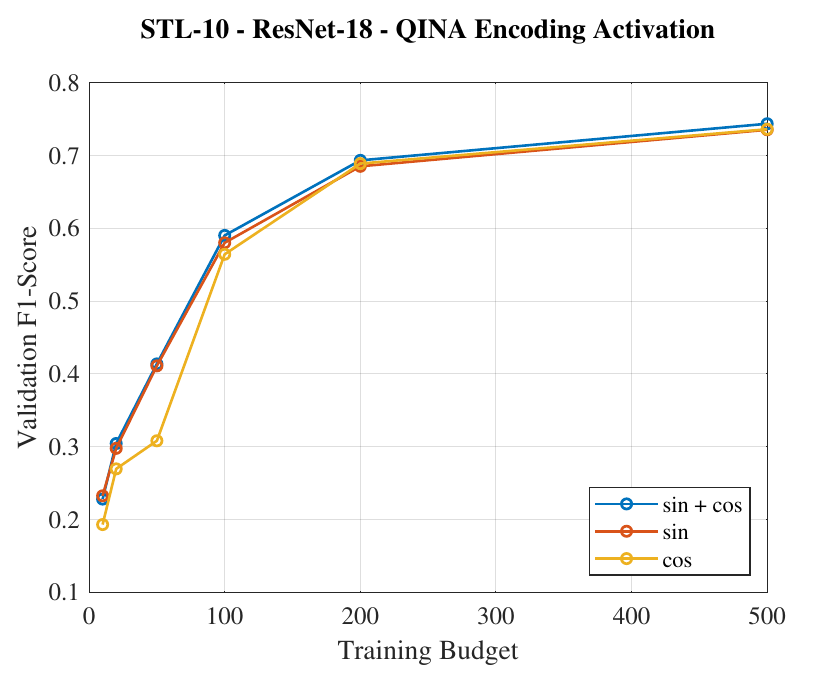}
\end{minipage}
\hfill
\begin{minipage}{0.49\textwidth}
\centering
%\caption*{B}
\includegraphics[width=\linewidth]{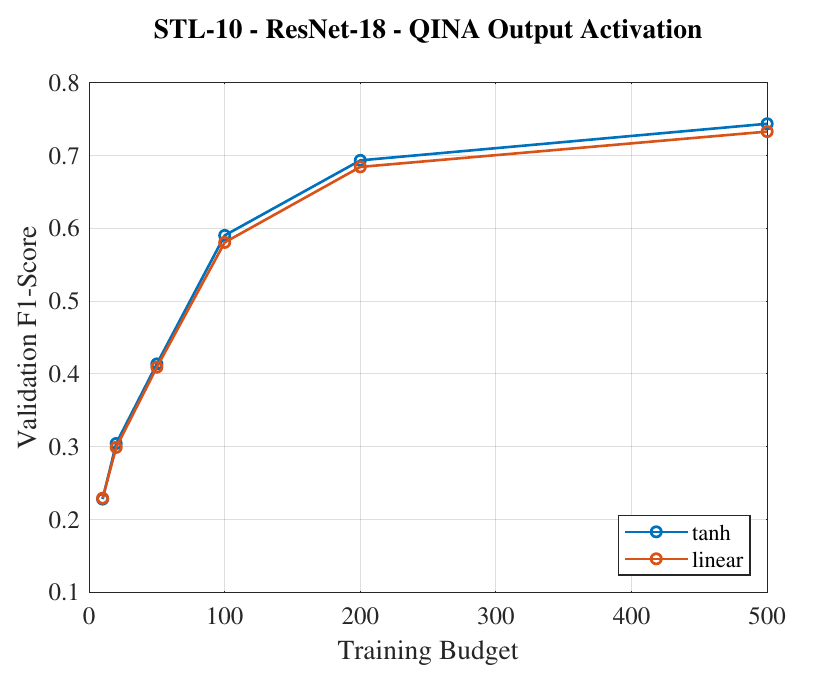}
\end{minipage}
\vspace{0.4cm}
\caption{QINA component ablations on STL-10 with ResNet-18. \textbf{Left:} sine-only, cosine-only, and combined sine--cosine encodings yield comparable performance, with the combined encoding providing a small advantage. \textbf{Right:} the bounded $\tanh$ output performs slightly better than a linear output across the evaluated training budgets.}
\label{fig:stl-10_class_activ_ablation}
\end{figure}

As shown in the left panel of \Cref{fig:stl-10_class_activ_ablation}, sine-only, cosine-only, and combined sine--cosine encodings achieve similar performance across training budgets. This behavior is expected because the three parameterizations are closely
related by phase shifts and scaling: $\cos(P)=\sin(P+\pi/2)$ and $\sin(P)+\cos(P)=\sqrt{2}\sin(P+\pi/4)$. Since the phase projection contains a learnable bias and is followed by a learned decoder, much of these differences can be absorbed during optimization. The combined sine--cosine encoding nevertheless provides a
small empirical advantage and is therefore used in QINA.

The right panel evaluates the final bounded nonlinearity by replacing $\tanh$ with a linear output. Both variants perform similarly, although $\tanh$ provides a small improvement across the evaluated budgets. This suggests that the main benefit does not depend critically on the output nonlinearity, while the bounded $\tanh$ transformation provides a modest empirical advantage and retains the bounded-output property used in the QINA formulation.

\section{Computational Complexity}

QINA introduces approximately $2Cd$ additional parameters through the encoder and decoder projections, corresponding to $\mathcal{O}(Cd)$ complexity, and incurs negligible runtime overhead relative to the backbone. In contrast, the PQC variant introduces additional computational cost due to classical simulation of the quantum circuit. State-vector simulation requires maintaining a $2^n$-dimensional quantum state, resulting in computational complexity proportional to $\mathcal{O}(L2^n)$ for $n$ qubits and circuit depth $L$. Replacing entire vision backbones with quantum circuits is therefore infeasible for realistic image resolutions, as the Hilbert space dimension grows exponentially with the number of qubits. Consequently, quantum components must be restricted to low-dimensional feature subspaces where the qubit count remains small.

\section{Adapter Configurations}

For reproducibility, we report the exact adapter configurations used for each backbone in \Cref{tab:adapter_config_resnet18,tab:adapter_config_resnet50,tab:adapter_config_xception,tab:adapter_config_vit}, including insertion location, projection dimensions, nonlinearities, and trainable parameter counts. Counts exclude the task-specific prediction head.

\begin{table}[H]
\centering
\caption{Adapter configurations used in the ResNet-18 comparisons. The adapter was applied to \textit{Residual Block 5a}. Weight dimensions are reported as input$\times$output; bias dimensions indicate the number of output units. Trainable parameter counts exclude the task head. For Fourier features, the randomly initialized encoder parameters are fixed during training.}
\label{tab:adapter_config_resnet18}
\vspace{0.25cm}
\footnotesize
\renewcommand{\arraystretch}{0.9}
\setlength{\tabcolsep}{4pt}
\begin{tabular}{llccc}
\toprule
Adapter & Component & Parameters & Activation & Learnable params \\
\midrule
\multirow{2}{*}{QINA}
& Encoder & $W:256\times128,\; b:128$ & $\sin+\cos$ & \multirow{2}{*}{65,920} \\
& Decoder & $W:128\times256,\; b:256$ & $\tanh$ & \\
\midrule
\multirow{2}{*}{MLP}
& Encoder & $W:256\times128,\; b:128$ & identity / ReLU / tanh & \multirow{2}{*}{65,920} \\
& Decoder & $W:128\times256,\; b:256$ & identity & \\
\midrule
\multirow{2}{*}{Fourier}
& Encoder & $W:256\times128,\; b:128$ (fixed) & $\sin$ & \multirow{2}{*}{33,024} \\
& Decoder & $W:128\times256,\; b:256$ & identity & \\
\midrule
\multirow{2}{*}{LoRA}
& Encoder & $W:256\times128$ & identity & \multirow{2}{*}{98,304} \\
& Decoder & $W:128\times512$ & identity & \\
\midrule
\multirow{4}{*}{Houlsby}
& Encoder 1 & $W:256\times64,\; b:64$ & GELU & \multirow{4}{*}{66,400} \\
& Decoder 1 & $W:64\times256,\; b:256$ & identity & \\
& Encoder 2 & $W:512\times32,\; b:32$ & GELU & \\
& Decoder 2 & $W:32\times512,\; b:512$ & identity & \\
\bottomrule
\end{tabular}
\end{table}

\begin{table}[H]
\small
\centering
\caption{Adapter configurations used in the ResNet-50 comparisons. The adapter was applied to \textit{Residual Block 5a}. Weight dimensions are reported as input$\times$output; bias dimensions indicate the number of output units. Trainable parameter counts exclude the task head. For Fourier features, the randomly initialized encoder parameters are fixed during training.}
\label{tab:adapter_config_resnet50}
\vspace{0.25cm}
\footnotesize
\renewcommand{\arraystretch}{0.9}
\setlength{\tabcolsep}{4pt}
\begin{tabular}{llccc}
\toprule
Adapter & Component & Parameters & Activation & Learnable params \\
\midrule
\multirow{2}{*}{QINA}
& Encoder & $W:1024\times128,\; b:128$ & $\sin+\cos$ & \multirow{2}{*}{263,296} \\
& Decoder & $W:128\times1024,\; b:1024$ & $\tanh$ & \\
\midrule
\multirow{2}{*}{MLP}
& Encoder & $W:1024\times128,\; b:128$ & identity / ReLU / tanh & \multirow{2}{*}{263,296} \\
& Decoder & $W:128\times1024,\; b:1024$ & identity & \\
\midrule
\multirow{2}{*}{Fourier}
& Encoder & $W:1024\times128,\; b:128$ (fixed) & $\sin$ & \multirow{2}{*}{132,096} \\
& Decoder & $W:128\times1024,\; b:1024$ & identity & \\
\midrule
\multirow{2}{*}{LoRA}
& Encoder & $W:1024\times128$ & identity & \multirow{2}{*}{393,216} \\
& Decoder & $W:128\times2048$ & identity & \\
\midrule
\multirow{4}{*}{Houlsby}
& Encoder 1 & $W:1024\times64,\; b:64$ & GELU & \multirow{4}{*}{265,312} \\
& Decoder 1 & $W:64\times1024,\; b:1024$ & identity & \\
& Encoder 2 & $W:2048\times32,\; b:32$ & GELU & \\
& Decoder 2 & $W:32\times2048,\; b:2048$ & identity & \\
\bottomrule
\end{tabular}
\end{table}

\begin{table}[H]
\centering
\caption{Adapter configurations used in the Xception comparisons. The adapter was applied to \textit{Block 13}. Weight dimensions are reported as input$\times$output; bias dimensions indicate the number of output units. Trainable parameter counts exclude the task head. For Fourier features, the randomly initialized encoder parameters are fixed during training.}
\label{tab:adapter_config_xception}
\vspace{0.25cm}
\footnotesize
\renewcommand{\arraystretch}{0.9}
\setlength{\tabcolsep}{4pt}
\begin{tabular}{llccc}
\toprule
Adapter & Component & Parameters & Activation & Learnable params \\
\midrule
\multirow{2}{*}{QINA}
& Encoder & $W:728\times128,\; b:128$ & $\sin+\cos$ & \multirow{2}{*}{187,224} \\
& Decoder & $W:128\times728,\; b:728$ & $\tanh$ & \\
\midrule
\multirow{2}{*}{MLP}
& Encoder & $W:728\times128,\; b:128$ & identity / ReLU / tanh & \multirow{2}{*}{187,224} \\
& Decoder & $W:128\times728,\; b:728$ & identity & \\
\midrule
\multirow{2}{*}{Fourier}
& Encoder & $W:728\times128,\; b:128$ (fixed) & $\sin$ & \multirow{2}{*}{93,912} \\
& Decoder & $W:128\times728,\; b:728$ & identity & \\
\midrule
\multirow{2}{*}{LoRA}
& Encoder & $W:728\times128$ & identity & \multirow{2}{*}{224,256} \\
& Decoder & $W:128\times1024$ & identity & \\
\midrule
\multirow{4}{*}{Houlsby}
& Encoder 1 & $W:728\times64,\; b:64$ & GELU & \multirow{4}{*}{189,254} \\
& Decoder 1 & $W:64\times728,\; b:728$ & identity & \\
& Encoder 2 & $W:1024\times46,\; b:46$ & GELU & \\
& Decoder 2 & $W:46\times1024,\; b:1024$ & identity & \\
\bottomrule
\end{tabular}
\end{table}

\begin{table}[H]
\centering
\caption{Adapter configurations used in the ViT-Tiny comparisons. The adapter was applied to \textit{Block 11}. Weight dimensions are reported as input$\times$output; bias dimensions indicate the number of output units. Trainable parameter counts exclude the task head. For Fourier features, the randomly initialized encoder parameters are fixed during training.}
\label{tab:adapter_config_vit}
\vspace{0.25cm}
\footnotesize
\renewcommand{\arraystretch}{0.9}
\setlength{\tabcolsep}{4pt}
\begin{tabular}{llccc}
\toprule
Adapter & Component & Parameters & Activation & Learnable params \\
\midrule
\multirow{2}{*}{QINA}
& Encoder & $W:192\times128,\; b:128$ & $\sin+\cos$ & \multirow{2}{*}{49,472} \\
& Decoder & $W:128\times192,\; b:192$ & $\tanh$ & \\
\midrule
\multirow{2}{*}{MLP}
& Encoder & $W:192\times128,\; b:128$ & identity / ReLU / tanh & \multirow{2}{*}{49,472} \\
& Decoder & $W:128\times192,\; b:192$ & identity & \\
\midrule
\multirow{2}{*}{Fourier}
& Encoder & $W:192\times128,\; b:128$ (fixed) & $\sin$ & \multirow{2}{*}{24,768} \\
& Decoder & $W:128\times192,\; b:192$ & identity & \\
\midrule
\multirow{2}{*}{LoRA}
& Encoder & $W:192\times128$ & identity & \multirow{2}{*}{49,152} \\
& Decoder & $W:128\times192$ & identity & \\
\midrule
\multirow{4}{*}{Houlsby}
& Encoder 1 & $W:192\times64,\; b:64$ & GELU & \multirow{4}{*}{49,664} \\
& Decoder 1 & $W:64\times192,\; b:192$ & identity & \\
& Encoder 2 & $W:192\times64,\; b:64$ & GELU & \\
& Decoder 2 & $W:64\times192,\; b:192$ & identity & \\
\bottomrule
\end{tabular}
\end{table}

\section{Quantitative Results}

This section presents the numerical results supporting the analyses discussed throughout the paper. We compare QINA against identity baselines, parameter-matched MLP adapters, and fixed Fourier feature mappings under varying budgets. All results report the mean$\pm$SD performance across five random seeds.

\subsection{Natural Image Classification}

We first evaluate the proposed adapter on STL-10 using multiple pretrained backbones. \Cref{tab:stl-10_class_models_f1,tab:stl-10_class_models_acc} report F1-score and accuracy across labeling budgets. These experiments assess the effect of structured nonlinear adaptation on in-domain natural image classification under frozen representation constraints.

\begin{table}[H]
\caption{Validation F1-score (mean$\pm$SD) on STL-10 under frozen-backbone training across different classification backbones and training budgets.}
\vspace{0.4cm}
\label{tab:stl-10_class_models_f1}
\centering
\renewcommand{\arraystretch}{0.9}
\resizebox{0.999\textwidth}{!}{
\begin{tabular}{lcccccc}
\toprule
\textbf{Training Budget} & \textbf{10} & \textbf{20} & \textbf{50} &  \textbf{100} & \textbf{200} & \textbf{500} \\
\midrule
& \multicolumn{6}{c}{\textbf{ResNet-18}} \\
\midrule
\textbf{QINA} & 0.228$\pm$0.054 & 0.304$\pm$0.048 & 0.414$\pm$0.039 & 0.590$\pm$0.025 & 0.693$\pm$0.040 & 0.744$\pm$0.022 \\
\textbf{Identity} & 0.114$\pm$0.027 & 0.116$\pm$0.026 & 0.124$\pm$0.018 & 0.272$\pm$0.033 & 0.396$\pm$0.047 & 0.402$\pm$0.034 \\
\textbf{Fourier} & 0.176$\pm$0.037 & 0.241$\pm$0.039 & 0.293$\pm$0.043 & 0.496$\pm$0.044 & 0.603$\pm$0.037 & 0.639$\pm$0.024 \\
\textbf{MLP-Linear} & 0.165$\pm$0.023 & 0.210$\pm$0.044 & 0.221$\pm$0.035 & 0.446$\pm$0.040 & 0.604$\pm$0.039 & 0.619$\pm$0.031 \\
\textbf{MLP-ReLU} & 0.159$\pm$0.027 & 0.190$\pm$0.042 & 0.199$\pm$0.029 & 0.417$\pm$0.038 & 0.586$\pm$0.041 & 0.607$\pm$0.034 \\
\textbf{MLP-Tanh} & 0.164$\pm$0.023 & 0.208$\pm$0.039 & 0.225$\pm$0.042 & 0.446$\pm$0.042 & 0.599$\pm$0.041 & 0.613$\pm$0.032 \\
\midrule
& \multicolumn{6}{c}{\textbf{Xception}} \\
\midrule
\textbf{QINA} & 0.313$\pm$0.093 & 0.469$\pm$0.110 & 0.512$\pm$0.065 & 0.844$\pm$0.025 & 0.923$\pm$0.021 & 0.952$\pm$0.020 \\
\textbf{Identity} & 0.262$\pm$0.054 & 0.368$\pm$0.076 & 0.381$\pm$0.044 & 0.755$\pm$0.040 & 0.822$\pm$0.051 & 0.859$\pm$0.030 \\
\textbf{Fourier} & 0.245$\pm$0.075 & 0.358$\pm$0.064 & 0.399$\pm$0.038 & 0.825$\pm$0.010 & 0.875$\pm$0.018 & 0.897$\pm$0.014 \\
\textbf{MLP-Linear} & 0.309$\pm$0.106 & 0.454$\pm$0.089 & 0.512$\pm$0.076 & 0.793$\pm$0.040 & 0.870$\pm$0.028 & 0.904$\pm$0.010 \\
\textbf{MLP-ReLU} & 0.328$\pm$0.116 & 0.453$\pm$0.077 & 0.479$\pm$0.058 & 0.796$\pm$0.043 & 0.869$\pm$0.029 & 0.900$\pm$0.013 \\
\textbf{MLP-Tanh} & 0.298$\pm$0.087 & 0.413$\pm$0.116 & 0.479$\pm$0.071 & 0.794$\pm$0.040 & 0.864$\pm$0.040 & 0.895$\pm$0.009 \\
\midrule
& \multicolumn{6}{c}{\textbf{ViT-Tiny}} \\
\midrule
\textbf{QINA} & 0.121$\pm$0.026 & 0.119$\pm$0.025 & 0.139$\pm$0.030 & 0.230$\pm$0.017 & 0.326$\pm$0.018 & 0.332$\pm$0.087 \\
\textbf{Identity} & 0.102$\pm$0.041 & 0.105$\pm$0.038 & 0.102$\pm$0.039 & 0.121$\pm$0.046 & 0.136$\pm$0.045 & 0.137$\pm$0.046 \\
\textbf{Fourier} & 0.115$\pm$0.031 & 0.125$\pm$0.037 & 0.132$\pm$0.045 & 0.199$\pm$0.037 & 0.240$\pm$0.039 & 0.258$\pm$0.026 \\
\textbf{MLP-Linear} & 0.111$\pm$0.016 & 0.116$\pm$0.015 & 0.123$\pm$0.007 & 0.177$\pm$0.016 & 0.226$\pm$0.017 & 0.225$\pm$0.023 \\
\textbf{MLP-ReLU} & 0.100$\pm$0.027 & 0.099$\pm$0.027 & 0.109$\pm$0.035 & 0.150$\pm$0.020 & 0.210$\pm$0.036 & 0.191$\pm$0.034 \\
\textbf{MLP-Tanh} & 0.119$\pm$0.021 & 0.119$\pm$0.004 & 0.126$\pm$0.012 & 0.186$\pm$0.026 & 0.227$\pm$0.025 & 0.233$\pm$0.030 \\
\bottomrule
\end{tabular}
}
\end{table}

\begin{table}[H]
\caption{Validation accuracy (mean$\pm$SD) on STL-10 under frozen-backbone training across different classification backbones and training budgets.}
\vspace{0.4cm}
\label{tab:stl-10_class_models_acc}
\centering
\renewcommand{\arraystretch}{0.9}
\resizebox{0.999\textwidth}{!}{
\begin{tabular}{lcccccc}
\toprule
\textbf{Training Budget} & \textbf{10} & \textbf{20} & \textbf{50} &  \textbf{100} & \textbf{200} & \textbf{500} \\
\midrule
& \multicolumn{6}{c}{\textbf{ResNet-18}} \\
\midrule
\textbf{QINA} & 0.248$\pm$0.052 & 0.328$\pm$0.056 & 0.431$\pm$0.032 & 0.604$\pm$0.029 & 0.703$\pm$0.037 & 0.753$\pm$0.021 \\
\textbf{Identity} & 0.152$\pm$0.044 & 0.151$\pm$0.049 & 0.155$\pm$0.035 & 0.303$\pm$0.034 & 0.418$\pm$0.039 & 0.425$\pm$0.031 \\
\textbf{Fourier} & 0.196$\pm$0.039 & 0.258$\pm$0.050 & 0.309$\pm$0.036 & 0.509$\pm$0.044 & 0.614$\pm$0.036 & 0.650$\pm$0.025 \\
\textbf{MLP-Linear} & 0.189$\pm$0.034 & 0.226$\pm$0.044 & 0.239$\pm$0.043 & 0.462$\pm$0.036 & 0.614$\pm$0.035 & 0.630$\pm$0.027 \\
\textbf{MLP-ReLU} & 0.178$\pm$0.036 & 0.209$\pm$0.040 & 0.220$\pm$0.042 & 0.436$\pm$0.035 & 0.595$\pm$0.037 & 0.617$\pm$0.033 \\
\textbf{MLP-Tanh} & 0.189$\pm$0.035 & 0.224$\pm$0.040 & 0.240$\pm$0.048 & 0.464$\pm$0.038 & 0.609$\pm$0.038 & 0.624$\pm$0.029 \\
\midrule
& \multicolumn{6}{c}{\textbf{Xception}} \\
\midrule
\textbf{QINA} & 0.346$\pm$0.074 & 0.500$\pm$0.084 & 0.549$\pm$0.036 & 0.846$\pm$0.022 & 0.923$\pm$0.020 & 0.950$\pm$0.022 \\
\textbf{Identity} & 0.299$\pm$0.049 & 0.394$\pm$0.060 & 0.409$\pm$0.036 & 0.758$\pm$0.032 & 0.825$\pm$0.044 & 0.860$\pm$0.028 \\
\textbf{Fourier} & 0.302$\pm$0.063 & 0.391$\pm$0.058 & 0.423$\pm$0.033 & 0.824$\pm$0.009 & 0.873$\pm$0.018 & 0.895$\pm$0.016 \\
\textbf{MLP-Linear} & 0.361$\pm$0.099 & 0.488$\pm$0.082 & 0.547$\pm$0.064 & 0.795$\pm$0.038 & 0.870$\pm$0.027 & 0.904$\pm$0.010 \\
\textbf{MLP-ReLU} & 0.375$\pm$0.109 & 0.485$\pm$0.073 & 0.520$\pm$0.060 & 0.798$\pm$0.040 & 0.870$\pm$0.029 & 0.900$\pm$0.013 \\
\textbf{MLP-Tanh} & 0.343$\pm$0.086 & 0.453$\pm$0.104 & 0.517$\pm$0.052 & 0.796$\pm$0.038 & 0.865$\pm$0.039 & 0.895$\pm$0.009 \\
\midrule
& \multicolumn{6}{c}{\textbf{ViT-Tiny}} \\
\midrule
\textbf{QINA} & 0.174$\pm$0.029 & 0.170$\pm$0.026 & 0.181$\pm$0.036 & 0.279$\pm$0.027 & 0.375$\pm$0.019 & 0.380$\pm$0.072 \\
\textbf{Identity} & 0.112$\pm$0.039 & 0.114$\pm$0.039 & 0.112$\pm$0.040 & 0.134$\pm$0.048 & 0.150$\pm$0.045 & 0.151$\pm$0.047 \\
\textbf{Fourier} & 0.137$\pm$0.031 & 0.153$\pm$0.040 & 0.158$\pm$0.049 & 0.229$\pm$0.048 & 0.278$\pm$0.051 & 0.292$\pm$0.031 \\
\textbf{MLP-Linear} & 0.134$\pm$0.035 & 0.142$\pm$0.029 & 0.149$\pm$0.027 & 0.209$\pm$0.029 & 0.260$\pm$0.036 & 0.257$\pm$0.042 \\
\textbf{MLP-ReLU} & 0.124$\pm$0.047 & 0.119$\pm$0.047 & 0.130$\pm$0.054 & 0.175$\pm$0.037 & 0.242$\pm$0.048 & 0.219$\pm$0.043 \\
\textbf{MLP-Tanh} & 0.142$\pm$0.043 & 0.144$\pm$0.026 & 0.152$\pm$0.032 & 0.215$\pm$0.037 & 0.262$\pm$0.042 & 0.265$\pm$0.047 \\
\bottomrule
\end{tabular}
}
\end{table}

\subsection{Medical Image Classification}

We next evaluate classification performance on medical imaging datasets using pretrained ResNet-18 backbones. \Cref{tab:medic_class_models_f1,tab:medic_class_models_acc} summarize F1-score and accuracy across training budgets and datasets. These experiments test whether the proposed adapter improves transfer from natural-image pretraining to medical imaging domains.

\subsection{Medical Image Segmentation}

Finally, we evaluate the adapters within DeepLabv3+ segmentation pipelines with pretrained ResNet-18 backbones. \Cref{tab:medic_seg_models_dice,tab:medic_seg_models_iou} report foreground Dice (F1-score) and Intersection-over-Union (IoU) across datasets and labeling budgets. Only the backbone representation is modified while the segmentation decoder remains unchanged, isolating the effect of the proposed adapters on frozen feature representations.

\begin{table}[H]
\caption{Validation F1-score (mean$\pm$SD) on medical datasets under frozen-ResNet-18 training across different training budgets.}
\vspace{0.4cm}
\label{tab:medic_class_models_f1}
\centering
\renewcommand{\arraystretch}{0.9}
\resizebox{0.999\textwidth}{!}{
\begin{tabular}{lcccccc}
\toprule
\textbf{Training Budget} & \textbf{10} & \textbf{20} & \textbf{50} &  \textbf{100} & \textbf{200} & \textbf{500} \\
\midrule
& \multicolumn{6}{c}{\textbf{BRISC}} \\
\midrule
\textbf{QINA} & 0.502$\pm$0.085 & 0.619$\pm$0.086 & 0.697$\pm$0.038 & 0.873$\pm$0.016 & 0.933$\pm$0.020 & 0.949$\pm$0.012 \\
\textbf{Identity} & 0.230$\pm$0.052 & 0.251$\pm$0.065 & 0.262$\pm$0.046 & 0.322$\pm$0.089 & 0.410$\pm$0.076 & 0.389$\pm$0.070 \\
\textbf{Fourier} & 0.412$\pm$0.073 & 0.465$\pm$0.095 & 0.508$\pm$0.081 & 0.728$\pm$0.041 & 0.811$\pm$0.034 & 0.826$\pm$0.022 \\
\textbf{MLP-Linear} & 0.418$\pm$0.060 & 0.506$\pm$0.073 & 0.529$\pm$0.073 & 0.759$\pm$0.029 & 0.831$\pm$0.025 & 0.834$\pm$0.021 \\
\textbf{MLP-ReLU} & 0.370$\pm$0.052 & 0.461$\pm$0.079 & 0.473$\pm$0.060 & 0.725$\pm$0.018 & 0.815$\pm$0.029 & 0.828$\pm$0.015 \\
\textbf{MLP-Tanh} & 0.407$\pm$0.060 & 0.508$\pm$0.072 & 0.509$\pm$0.072 & 0.752$\pm$0.031 & 0.828$\pm$0.018 & 0.834$\pm$0.025 \\
\midrule
& \multicolumn{6}{c}{\textbf{ISIC}} \\
\midrule
\textbf{QINA} & 0.151$\pm$0.020 & 0.183$\pm$0.013 & 0.208$\pm$0.022 & 0.239$\pm$0.019 & 0.264$\pm$0.011 & 0.296$\pm$0.011 \\
\textbf{Identity} & 0.116$\pm$0.025 & 0.112$\pm$0.023 & 0.112$\pm$0.024 & 0.145$\pm$0.023 & 0.165$\pm$0.025 & 0.169$\pm$0.018 \\
\textbf{Fourier} & 0.133$\pm$0.020 & 0.153$\pm$0.007 & 0.162$\pm$0.014 & 0.206$\pm$0.016 & 0.231$\pm$0.010 & 0.244$\pm$0.011 \\
\textbf{MLP-Linear} & 0.133$\pm$0.018 & 0.147$\pm$0.014 & 0.150$\pm$0.014 & 0.194$\pm$0.015 & 0.222$\pm$0.016 & 0.244$\pm$0.003 \\
\textbf{MLP-ReLU} & 0.131$\pm$0.019 & 0.139$\pm$0.006 & 0.145$\pm$0.015 & 0.193$\pm$0.019 & 0.223$\pm$0.013 & 0.237$\pm$0.009 \\
\textbf{MLP-Tanh} & 0.131$\pm$0.017 & 0.150$\pm$0.014 & 0.150$\pm$0.016 & 0.194$\pm$0.016 & 0.222$\pm$0.017 & 0.244$\pm$0.005 \\
\midrule
& \multicolumn{6}{c}{\textbf{FETAL}} \\
\midrule
\textbf{QINA} & 0.412$\pm$0.039 & 0.499$\pm$0.052 & 0.547$\pm$0.043 & 0.645$\pm$0.026 & 0.736$\pm$0.034 & 0.765$\pm$0.025 \\
\textbf{Identity} & 0.168$\pm$0.030 & 0.209$\pm$0.061 & 0.199$\pm$0.054 & 0.309$\pm$0.051 & 0.397$\pm$0.028 & 0.389$\pm$0.028 \\
\textbf{Fourier} & 0.334$\pm$0.060 & 0.423$\pm$0.029 & 0.450$\pm$0.055 & 0.545$\pm$0.013 & 0.611$\pm$0.034 & 0.603$\pm$0.016 \\
\textbf{MLP-Linear} & 0.305$\pm$0.053 & 0.365$\pm$0.024 & 0.384$\pm$0.028 & 0.507$\pm$0.028 & 0.554$\pm$0.036 & 0.550$\pm$0.038 \\
\textbf{MLP-ReLU} & 0.287$\pm$0.063 & 0.361$\pm$0.024 & 0.371$\pm$0.026 & 0.494$\pm$0.038 & 0.535$\pm$0.045 & 0.537$\pm$0.045 \\
\textbf{MLP-Tanh} & 0.302$\pm$0.056 & 0.363$\pm$0.022 & 0.386$\pm$0.031 & 0.506$\pm$0.023 & 0.555$\pm$0.042 & 0.554$\pm$0.038 \\
\midrule
& \multicolumn{6}{c}{\textbf{BUSI}} \\
\midrule
\textbf{QINA} & 0.345$\pm$0.066 & 0.361$\pm$0.078 & 0.402$\pm$0.079 & 0.524$\pm$0.104 & 0.637$\pm$0.101 & 0.642$\pm$0.099 \\
\textbf{Identity} & 0.282$\pm$0.042 & 0.257$\pm$0.024 & 0.267$\pm$0.024 & 0.294$\pm$0.050 & 0.298$\pm$0.052 & 0.295$\pm$0.056 \\
\textbf{Fourier} & 0.310$\pm$0.065 & 0.330$\pm$0.076 & 0.329$\pm$0.079 & 0.411$\pm$0.082 & 0.427$\pm$0.074 & 0.447$\pm$0.079 \\
\textbf{MLP-Linear} & 0.338$\pm$0.070 & 0.347$\pm$0.089 & 0.365$\pm$0.083 & 0.426$\pm$0.084 & 0.491$\pm$0.074 & 0.494$\pm$0.083 \\
\textbf{MLP-ReLU} & 0.307$\pm$0.046 & 0.320$\pm$0.061 & 0.329$\pm$0.070 & 0.391$\pm$0.083 & 0.467$\pm$0.060 & 0.489$\pm$0.098 \\
\textbf{MLP-Tanh} & 0.340$\pm$0.074 & 0.337$\pm$0.075 & 0.366$\pm$0.078 & 0.424$\pm$0.066 & 0.485$\pm$0.076 & 0.495$\pm$0.087 \\
\midrule
& \multicolumn{6}{c}{\textbf{BUSI (FETAL-Adapted)}} \\
\midrule
\textbf{QINA} & 0.441$\pm$0.062 & 0.472$\pm$0.058 & 0.590$\pm$0.042 & 0.732$\pm$0.047 & 0.786$\pm$0.046 & 0.852$\pm$0.033 \\
\textbf{Identity} & 0.311$\pm$0.044 & 0.297$\pm$0.043 & 0.307$\pm$0.045 & 0.356$\pm$0.055 & 0.423$\pm$0.046 & 0.404$\pm$0.030 \\
\textbf{Fourier} & 0.397$\pm$0.052 & 0.396$\pm$0.066 & 0.469$\pm$0.056 & 0.625$\pm$0.069 & 0.697$\pm$0.040 & 0.742$\pm$0.032 \\
\textbf{MLP-Linear} & 0.402$\pm$0.040 & 0.416$\pm$0.061 & 0.441$\pm$0.039 & 0.632$\pm$0.058 & 0.705$\pm$0.047 & 0.731$\pm$0.029 \\
\textbf{MLP-ReLU} & 0.388$\pm$0.012 & 0.398$\pm$0.053 & 0.423$\pm$0.016 & 0.612$\pm$0.057 & 0.681$\pm$0.042 & 0.718$\pm$0.013 \\
\textbf{MLP-Tanh} & 0.392$\pm$0.036 & 0.421$\pm$0.069 & 0.436$\pm$0.036 & 0.636$\pm$0.064 & 0.704$\pm$0.049 & 0.734$\pm$0.032 \\
\bottomrule
\end{tabular}
}
\end{table}

\begin{table}[H]
\caption{Validation accuracy (mean$\pm$SD) on medical datasets under frozen-ResNet-18 training across different training budgets.}
\vspace{0.4cm}
\label{tab:medic_class_models_acc}
\centering
\renewcommand{\arraystretch}{0.9}
\resizebox{0.999\textwidth}{!}{
\begin{tabular}{lcccccc}
\toprule
\textbf{Training Budget} & \textbf{10} & \textbf{20} & \textbf{50} &  \textbf{100} & \textbf{200} & \textbf{500} \\
\midrule
& \multicolumn{6}{c}{\textbf{BRISC}} \\
\midrule
\textbf{QINA} & 0.534$\pm$0.072 & 0.637$\pm$0.077 & 0.714$\pm$0.033 & 0.876$\pm$0.012 & 0.933$\pm$0.020 & 0.949$\pm$0.011 \\
\textbf{Identity} & 0.278$\pm$0.047 & 0.296$\pm$0.047 & 0.303$\pm$0.042 & 0.354$\pm$0.073 & 0.439$\pm$0.061 & 0.423$\pm$0.052 \\
\textbf{Fourier} & 0.449$\pm$0.051 & 0.500$\pm$0.081 & 0.545$\pm$0.064 & 0.736$\pm$0.039 & 0.814$\pm$0.036 & 0.829$\pm$0.023 \\
\textbf{MLP-Linear} & 0.450$\pm$0.072 & 0.524$\pm$0.053 & 0.548$\pm$0.059 & 0.763$\pm$0.027 & 0.833$\pm$0.024 & 0.837$\pm$0.019 \\
\textbf{MLP-ReLU} & 0.409$\pm$0.061 & 0.488$\pm$0.053 & 0.507$\pm$0.050 & 0.730$\pm$0.020 & 0.818$\pm$0.027 & 0.829$\pm$0.016 \\
\textbf{MLP-Tanh} & 0.442$\pm$0.073 & 0.524$\pm$0.055 & 0.534$\pm$0.057 & 0.757$\pm$0.030 & 0.829$\pm$0.018 & 0.835$\pm$0.024 \\
\midrule
& \multicolumn{6}{c}{\textbf{ISIC}} \\
\midrule
\textbf{QINA} & 0.342$\pm$0.032 & 0.405$\pm$0.021 & 0.440$\pm$0.020 & 0.509$\pm$0.023 & 0.548$\pm$0.019 & 0.592$\pm$0.010 \\
\textbf{Identity} & 0.256$\pm$0.050 & 0.278$\pm$0.030 & 0.271$\pm$0.034 & 0.335$\pm$0.049 & 0.394$\pm$0.040 & 0.394$\pm$0.029 \\
\textbf{Fourier} & 0.301$\pm$0.027 & 0.344$\pm$0.020 & 0.352$\pm$0.013 & 0.436$\pm$0.015 & 0.487$\pm$0.023 & 0.515$\pm$0.007 \\
\textbf{MLP-Linear} & 0.302$\pm$0.034 & 0.338$\pm$0.014 & 0.347$\pm$0.021 & 0.431$\pm$0.027 & 0.485$\pm$0.022 & 0.509$\pm$0.012 \\
\textbf{MLP-ReLU} & 0.305$\pm$0.052 & 0.326$\pm$0.016 & 0.339$\pm$0.028 & 0.426$\pm$0.035 & 0.480$\pm$0.025 & 0.494$\pm$0.023 \\
\textbf{MLP-Tanh} & 0.297$\pm$0.032 & 0.340$\pm$0.019 & 0.351$\pm$0.028 & 0.429$\pm$0.030 & 0.483$\pm$0.024 & 0.509$\pm$0.014 \\
\midrule
& \multicolumn{6}{c}{\textbf{FETAL}} \\
\midrule
\textbf{QINA} & 0.580$\pm$0.066 & 0.673$\pm$0.046 & 0.732$\pm$0.039 & 0.816$\pm$0.027 & 0.865$\pm$0.018 & 0.886$\pm$0.013 \\
\textbf{Identity} & 0.348$\pm$0.049 & 0.388$\pm$0.075 & 0.379$\pm$0.066 & 0.492$\pm$0.063 & 0.587$\pm$0.038 & 0.578$\pm$0.034 \\
\textbf{Fourier} & 0.492$\pm$0.062 & 0.593$\pm$0.023 & 0.632$\pm$0.045 & 0.719$\pm$0.028 & 0.759$\pm$0.020 & 0.762$\pm$0.014 \\
\textbf{MLP-Linear} & 0.462$\pm$0.049 & 0.539$\pm$0.026 & 0.573$\pm$0.047 & 0.697$\pm$0.013 & 0.726$\pm$0.015 & 0.728$\pm$0.017 \\
\textbf{MLP-ReLU} & 0.453$\pm$0.067 & 0.530$\pm$0.025 & 0.566$\pm$0.046 & 0.689$\pm$0.013 & 0.721$\pm$0.017 & 0.723$\pm$0.014 \\
\textbf{MLP-Tanh} & 0.460$\pm$0.054 & 0.536$\pm$0.024 & 0.575$\pm$0.049 & 0.695$\pm$0.008 & 0.729$\pm$0.017 & 0.732$\pm$0.016 \\
\midrule
& \multicolumn{6}{c}{\textbf{BUSI}} \\
\midrule
\textbf{QINA} & 0.561$\pm$0.064 & 0.577$\pm$0.021 & 0.601$\pm$0.036 & 0.688$\pm$0.042 & 0.750$\pm$0.050 & 0.756$\pm$0.053 \\
\textbf{Identity} & 0.513$\pm$0.054 & 0.490$\pm$0.080 & 0.509$\pm$0.059 & 0.533$\pm$0.028 & 0.551$\pm$0.008 & 0.542$\pm$0.023 \\
\textbf{Fourier} & 0.526$\pm$0.042 & 0.531$\pm$0.024 & 0.536$\pm$0.027 & 0.596$\pm$0.025 & 0.605$\pm$0.035 & 0.622$\pm$0.040 \\
\textbf{MLP-Linear} & 0.462$\pm$0.109 & 0.474$\pm$0.101 & 0.537$\pm$0.065 & 0.551$\pm$0.064 & 0.605$\pm$0.069 & 0.600$\pm$0.087 \\
\textbf{MLP-ReLU} & 0.444$\pm$0.129 & 0.454$\pm$0.112 & 0.515$\pm$0.070 & 0.538$\pm$0.056 & 0.595$\pm$0.063 & 0.610$\pm$0.074 \\
\textbf{MLP-Tanh} & 0.463$\pm$0.122 & 0.464$\pm$0.097 & 0.536$\pm$0.056 & 0.542$\pm$0.055 & 0.597$\pm$0.067 & 0.600$\pm$0.094 \\
\midrule
& \multicolumn{6}{c}{\textbf{BUSI (FETAL-Adapted)}} \\
\midrule
\textbf{QINA} & 0.634$\pm$0.031 & 0.662$\pm$0.037 & 0.727$\pm$0.028 & 0.811$\pm$0.031 & 0.840$\pm$0.031 & 0.882$\pm$0.027 \\
\textbf{Identity} & 0.535$\pm$0.029 & 0.545$\pm$0.037 & 0.544$\pm$0.046 & 0.590$\pm$0.018 & 0.629$\pm$0.021 & 0.617$\pm$0.017 \\
\textbf{Fourier} & 0.587$\pm$0.014 & 0.604$\pm$0.027 & 0.636$\pm$0.037 & 0.726$\pm$0.045 & 0.759$\pm$0.025 & 0.788$\pm$0.024 \\
\textbf{MLP-Linear} & 0.558$\pm$0.051 & 0.600$\pm$0.056 & 0.627$\pm$0.032 & 0.729$\pm$0.032 & 0.768$\pm$0.032 & 0.778$\pm$0.025 \\
\textbf{MLP-ReLU} & 0.559$\pm$0.060 & 0.603$\pm$0.058 & 0.618$\pm$0.022 & 0.717$\pm$0.027 & 0.751$\pm$0.023 & 0.768$\pm$0.018 \\
\textbf{MLP-Tanh} & 0.555$\pm$0.051 & 0.606$\pm$0.061 & 0.626$\pm$0.032 & 0.729$\pm$0.033 & 0.765$\pm$0.033 & 0.779$\pm$0.027 \\
\bottomrule
\end{tabular}
}
\end{table}

\begin{table}[H]
\caption{Validation Dice (mean$\pm$SD) on medical datasets using DeepLabv3+ with a frozen ResNet-18 across different training budgets.}
\vspace{0.4cm}
\label{tab:medic_seg_models_dice}
\centering
\renewcommand{\arraystretch}{0.9}
\resizebox{0.999\textwidth}{!}{
\begin{tabular}{lccccccc}
\toprule
\textbf{Training Budget} & \textbf{5} & \textbf{10} & \textbf{20} & \textbf{50} &  \textbf{100} & \textbf{200} & \textbf{500} \\
\midrule
& \multicolumn{7}{c}{\textbf{BRISC}} \\
\midrule
\textbf{QINA} & 0.021$\pm$0.015 & 0.082$\pm$0.054 & 0.168$\pm$0.079 & 0.400$\pm$0.043 & 0.405$\pm$0.069 & 0.492$\pm$0.105 & 0.533$\pm$0.072 \\
\textbf{Identity} & 0.012$\pm$0.009 & 0.074$\pm$0.052 & 0.147$\pm$0.058 & 0.330$\pm$0.036 & 0.375$\pm$0.026 & 0.435$\pm$0.052 & 0.435$\pm$0.064 \\
\textbf{Fourier} & 0.020$\pm$0.013 & 0.071$\pm$0.050 & 0.169$\pm$0.063 & 0.321$\pm$0.031 & 0.341$\pm$0.063 & 0.436$\pm$0.104 & 0.485$\pm$0.049 \\
\textbf{MLP-Linear} & 0.025$\pm$0.025 & 0.090$\pm$0.055 & 0.202$\pm$0.041 & 0.275$\pm$0.091 & 0.348$\pm$0.046 & 0.467$\pm$0.064 & 0.434$\pm$0.102 \\
\textbf{MLP-ReLU} & 0.019$\pm$0.018 & 0.088$\pm$0.042 & 0.182$\pm$0.057 & 0.254$\pm$0.071 & 0.362$\pm$0.048 & 0.446$\pm$0.063 & 0.413$\pm$0.102 \\
\textbf{MLP-Tanh} & 0.023$\pm$0.023 & 0.092$\pm$0.059 & 0.210$\pm$0.053 & 0.278$\pm$0.093 & 0.348$\pm$0.044 & 0.473$\pm$0.050 & 0.429$\pm$0.111 \\
\midrule
& \multicolumn{7}{c}{\textbf{BUSI}} \\
\midrule
\textbf{QINA} & 0.159$\pm$0.049 & 0.198$\pm$0.041 & 0.252$\pm$0.093 & 0.463$\pm$0.090 & 0.469$\pm$0.077 & 0.465$\pm$0.032 & 0.470$\pm$0.113 \\
\textbf{Identity} & 0.151$\pm$0.058 & 0.167$\pm$0.032 & 0.241$\pm$0.086 & 0.377$\pm$0.088 & 0.377$\pm$0.095 & 0.442$\pm$0.041 & 0.372$\pm$0.131 \\
\textbf{Fourier} & 0.162$\pm$0.059 & 0.184$\pm$0.031 & 0.236$\pm$0.102 & 0.402$\pm$0.090 & 0.410$\pm$0.077 & 0.426$\pm$0.043 & 0.393$\pm$0.103 \\
\textbf{MLP-Linear} & 0.136$\pm$0.028 & 0.161$\pm$0.025 & 0.290$\pm$0.015 & 0.387$\pm$0.097 & 0.375$\pm$0.068 & 0.347$\pm$0.036 & 0.404$\pm$0.091 \\
\textbf{MLP-ReLU} & 0.143$\pm$0.042 & 0.148$\pm$0.040 & 0.285$\pm$0.020 & 0.371$\pm$0.101 & 0.357$\pm$0.074 & 0.349$\pm$0.039 & 0.413$\pm$0.067 \\
\textbf{MLP-Tanh} & 0.136$\pm$0.028 & 0.159$\pm$0.024 & 0.288$\pm$0.017 & 0.385$\pm$0.102 & 0.372$\pm$0.068 & 0.345$\pm$0.031 & 0.398$\pm$0.094 \\
\midrule
& \multicolumn{7}{c}{\textbf{Kvasir}} \\
\midrule
\textbf{QINA} & 0.404$\pm$0.130 & 0.492$\pm$0.054 & 0.580$\pm$0.032 & 0.680$\pm$0.005 & 0.705$\pm$0.017 & 0.729$\pm$0.018 & 0.772$\pm$0.013 \\
\textbf{Identity} & 0.371$\pm$0.111 & 0.413$\pm$0.069 & 0.503$\pm$0.030 & 0.559$\pm$0.010 & 0.579$\pm$0.032 & 0.600$\pm$0.027 & 0.631$\pm$0.037 \\
\textbf{Fourier} & 0.379$\pm$0.119 & 0.446$\pm$0.054 & 0.521$\pm$0.033 & 0.607$\pm$0.005 & 0.634$\pm$0.016 & 0.649$\pm$0.025 & 0.691$\pm$0.012 \\
\textbf{MLP-Linear} & 0.367$\pm$0.117 & 0.455$\pm$0.026 & 0.540$\pm$0.033 & 0.594$\pm$0.017 & 0.632$\pm$0.015 & 0.655$\pm$0.025 & 0.670$\pm$0.018 \\
\textbf{MLP-ReLU} & 0.373$\pm$0.118 & 0.451$\pm$0.037 & 0.531$\pm$0.035 & 0.594$\pm$0.013 & 0.633$\pm$0.021 & 0.648$\pm$0.027 & 0.665$\pm$0.025 \\
\textbf{MLP-Tanh} & 0.367$\pm$0.116 & 0.454$\pm$0.026 & 0.538$\pm$0.035 & 0.593$\pm$0.017 & 0.631$\pm$0.015 & 0.654$\pm$0.025 & 0.668$\pm$0.019 \\
\midrule
& \multicolumn{7}{c}{\textbf{Cardiac}} \\
\midrule
\textbf{QINA} & 0.270$\pm$0.198 & 0.391$\pm$0.059 & 0.647$\pm$0.095 & 0.803$\pm$0.048 & 0.856$\pm$0.039 & 0.884$\pm$0.020 & 0.919$\pm$0.027 \\
\textbf{Identity} & 0.206$\pm$0.119 & 0.311$\pm$0.108 & 0.493$\pm$0.085 & 0.707$\pm$0.045 & 0.742$\pm$0.042 & 0.772$\pm$0.029 & 0.829$\pm$0.020 \\
\textbf{Fourier} & 0.228$\pm$0.179 & 0.352$\pm$0.077 & 0.606$\pm$0.054 & 0.735$\pm$0.043 & 0.763$\pm$0.061 & 0.793$\pm$0.030 & 0.839$\pm$0.026 \\
\textbf{MLP-Linear} & 0.189$\pm$0.131 & 0.425$\pm$0.169 & 0.504$\pm$0.101 & 0.686$\pm$0.075 & 0.782$\pm$0.028 & 0.802$\pm$0.023 & 0.826$\pm$0.033 \\
\textbf{MLP-ReLU} & 0.184$\pm$0.153 & 0.393$\pm$0.190 & 0.506$\pm$0.108 & 0.664$\pm$0.103 & 0.774$\pm$0.042 & 0.795$\pm$0.025 & 0.800$\pm$0.041 \\
\textbf{MLP-Tanh} & 0.184$\pm$0.126 & 0.421$\pm$0.169 & 0.509$\pm$0.101 & 0.682$\pm$0.076 & 0.787$\pm$0.028 & 0.801$\pm$0.024 & 0.823$\pm$0.033 \\
\bottomrule
\end{tabular}
}
\end{table}

\begin{table}[H]
\caption{Validation IoU (mean$\pm$SD) on medical datasets using DeepLabv3+ with a frozen ResNet-18 across different training budgets.}
\vspace{0.4cm}
\label{tab:medic_seg_models_iou}
\centering
\renewcommand{\arraystretch}{0.9}
\resizebox{0.999\textwidth}{!}{
\begin{tabular}{lccccccc}
\toprule
\textbf{Training Budget} & \textbf{5} & \textbf{10} & \textbf{20} & \textbf{50} &  \textbf{100} & \textbf{200} & \textbf{500} \\
\midrule
& \multicolumn{7}{c}{\textbf{BRISC}} \\
\midrule
\textbf{QINA} & 0.013$\pm$0.009 & 0.055$\pm$0.036 & 0.120$\pm$0.060 & 0.304$\pm$0.033 & 0.309$\pm$0.054 & 0.395$\pm$0.095 & 0.423$\pm$0.061 \\
\textbf{Identity} & 0.008$\pm$0.006 & 0.049$\pm$0.034 & 0.105$\pm$0.044 & 0.250$\pm$0.029 & 0.290$\pm$0.023 & 0.341$\pm$0.047 & 0.341$\pm$0.054 \\
\textbf{Fourier} & 0.012$\pm$0.008 & 0.047$\pm$0.033 & 0.119$\pm$0.048 & 0.239$\pm$0.029 & 0.259$\pm$0.049 & 0.346$\pm$0.092 & 0.381$\pm$0.039 \\
\textbf{MLP-Linear} & 0.017$\pm$0.017 & 0.061$\pm$0.038 & 0.146$\pm$0.031 & 0.205$\pm$0.069 & 0.264$\pm$0.036 & 0.369$\pm$0.054 & 0.343$\pm$0.087 \\
\textbf{MLP-ReLU} & 0.012$\pm$0.012 & 0.060$\pm$0.029 & 0.131$\pm$0.043 & 0.188$\pm$0.054 & 0.276$\pm$0.041 & 0.351$\pm$0.055 & 0.326$\pm$0.086 \\
\textbf{MLP-Tanh} & 0.015$\pm$0.015 & 0.062$\pm$0.041 & 0.151$\pm$0.040 & 0.207$\pm$0.071 & 0.264$\pm$0.036 & 0.373$\pm$0.042 & 0.338$\pm$0.095 \\
\midrule
& \multicolumn{7}{c}{\textbf{BUSI}} \\
\midrule
\textbf{QINA} & 0.106$\pm$0.032 & 0.136$\pm$0.030 & 0.183$\pm$0.068 & 0.348$\pm$0.071 & 0.353$\pm$0.060 & 0.362$\pm$0.029 & 0.367$\pm$0.094 \\
\textbf{Identity} & 0.099$\pm$0.040 & 0.117$\pm$0.026 & 0.170$\pm$0.061 & 0.285$\pm$0.069 & 0.280$\pm$0.073 & 0.338$\pm$0.034 & 0.284$\pm$0.105 \\
\textbf{Fourier} & 0.109$\pm$0.039 & 0.125$\pm$0.019 & 0.168$\pm$0.073 & 0.299$\pm$0.070 & 0.306$\pm$0.060 & 0.327$\pm$0.035 & 0.302$\pm$0.083 \\
\textbf{MLP-Linear} & 0.089$\pm$0.019 & 0.110$\pm$0.023 & 0.206$\pm$0.009 & 0.289$\pm$0.080 & 0.280$\pm$0.058 & 0.260$\pm$0.032 & 0.309$\pm$0.078 \\
\textbf{MLP-ReLU} & 0.093$\pm$0.029 & 0.101$\pm$0.031 & 0.202$\pm$0.015 & 0.275$\pm$0.083 & 0.265$\pm$0.060 & 0.260$\pm$0.031 & 0.314$\pm$0.057 \\
\textbf{MLP-Tanh} & 0.089$\pm$0.019 & 0.108$\pm$0.022 & 0.204$\pm$0.012 & 0.287$\pm$0.085 & 0.277$\pm$0.058 & 0.258$\pm$0.030 & 0.304$\pm$0.080 \\
\midrule
& \multicolumn{7}{c}{\textbf{Kvasir}} \\
\midrule
\textbf{QINA} & 0.292$\pm$0.098 & 0.361$\pm$0.046 & 0.443$\pm$0.028 & 0.546$\pm$0.006 & 0.576$\pm$0.016 & 0.603$\pm$0.021 & 0.654$\pm$0.013 \\
\textbf{Identity} & 0.262$\pm$0.085 & 0.295$\pm$0.057 & 0.374$\pm$0.025 & 0.431$\pm$0.009 & 0.454$\pm$0.027 & 0.477$\pm$0.024 & 0.509$\pm$0.032 \\
\textbf{Fourier} & 0.272$\pm$0.090 & 0.324$\pm$0.045 & 0.394$\pm$0.030 & 0.482$\pm$0.006 & 0.512$\pm$0.014 & 0.527$\pm$0.027 & 0.577$\pm$0.012 \\
\textbf{MLP-Linear} & 0.261$\pm$0.091 & 0.332$\pm$0.021 & 0.409$\pm$0.030 & 0.469$\pm$0.015 & 0.508$\pm$0.013 & 0.536$\pm$0.021 & 0.552$\pm$0.019 \\
\textbf{MLP-ReLU} & 0.266$\pm$0.092 & 0.329$\pm$0.031 & 0.401$\pm$0.032 & 0.468$\pm$0.012 & 0.509$\pm$0.019 & 0.528$\pm$0.023 & 0.547$\pm$0.025 \\
\textbf{MLP-Tanh} & 0.261$\pm$0.091 & 0.332$\pm$0.021 & 0.407$\pm$0.032 & 0.468$\pm$0.015 & 0.507$\pm$0.013 & 0.535$\pm$0.021 & 0.551$\pm$0.021 \\
\midrule
& \multicolumn{7}{c}{\textbf{Cardiac}} \\
\midrule
\textbf{QINA} & 0.204$\pm$0.155 & 0.286$\pm$0.043 & 0.520$\pm$0.092 & 0.685$\pm$0.050 & 0.742$\pm$0.043 & 0.777$\pm$0.025 & 0.823$\pm$0.033 \\
\textbf{Identity} & 0.150$\pm$0.090 & 0.228$\pm$0.083 & 0.384$\pm$0.071 & 0.592$\pm$0.049 & 0.635$\pm$0.041 & 0.666$\pm$0.030 & 0.735$\pm$0.024 \\
\textbf{Fourier} & 0.170$\pm$0.138 & 0.257$\pm$0.055 & 0.484$\pm$0.056 & 0.623$\pm$0.043 & 0.656$\pm$0.066 & 0.690$\pm$0.034 & 0.750$\pm$0.030 \\
\textbf{MLP-Linear} & 0.135$\pm$0.100 & 0.322$\pm$0.138 & 0.391$\pm$0.092 & 0.570$\pm$0.073 & 0.674$\pm$0.033 & 0.700$\pm$0.023 & 0.730$\pm$0.040 \\
\textbf{MLP-ReLU} & 0.133$\pm$0.117 & 0.298$\pm$0.152 & 0.391$\pm$0.100 & 0.548$\pm$0.104 & 0.668$\pm$0.045 & 0.691$\pm$0.026 & 0.702$\pm$0.047 \\
\textbf{MLP-Tanh} & 0.131$\pm$0.095 & 0.319$\pm$0.138 & 0.395$\pm$0.092 & 0.564$\pm$0.075 & 0.680$\pm$0.033 & 0.699$\pm$0.024 & 0.727$\pm$0.040 \\
\bottomrule
\end{tabular}
}
\end{table}

\end{appendices}

\end{document}